\documentclass{article}

   \usepackage[preprint,nonatbib]{neurips_2021}
\usepackage[utf8]{inputenc} 
\usepackage[T1]{fontenc}    
\usepackage{hyperref}       
\usepackage{url}            
\usepackage{booktabs}       
\usepackage{amsfonts}       
\usepackage{nicefrac}       
\usepackage{microtype}      
\usepackage{xcolor}         
\usepackage{graphicx}
\usepackage{todonotes}
\usepackage{siunitx}
\usepackage{multirow}
\usepackage{amssymb}
\usepackage{caption} 
\usepackage{subfig}
\usepackage{amsmath}
\usepackage[ruled,boxed,lined]{algorithm2e}
\usepackage{hyperref}
\hypersetup{
    colorlinks=true,
    linkcolor=blue,
    filecolor=magenta,      
    urlcolor=blue,
    citecolor=blue,
    pdfpagemode=FullScreen,
    }
\newcounter{todocnt}

\title{Multitask Adaptation by Retrospective Exploration with Learned World Models}

\author{%
  Artem Zholus \\
  Moscow Institute of Physics and Technology \\ Moscow, Russia \\ \texttt{zholus.aa@phystech.edu} \\
   \And
   Aleksandr I. Panov \\
   Artificial Intelligence Research Institute \\ Moscow, Russia \\ \texttt{panov.ai@ya.ru} \\
}
\begin{document}

\maketitle

\begin{abstract}
Model-based reinforcement learning (MBRL) allows solving complex tasks in a sample-efficient manner. However, no information is reused between the tasks. In this work, we propose a meta-learned addressing model called RAMa that provides training samples for the MBRL agent taken from continuously growing task-agnostic storage. The model is trained to maximize the expected agent's performance by selecting promising trajectories solving prior tasks from the storage. We show that such retrospective exploration can accelerate the learning process of the MBRL agent by better informing learned dynamics and prompting agent with exploratory trajectories. We test the performance of our approach on several domains from the DeepMind control suite, from Metaworld multitask benchmark, and from our bespoke environment implemented with a robotic NVIDIA Isaac simulator to test the ability of the model to act in a photorealistic, ray-traced environment.
\end{abstract}

\section{Introduction}

We address the problem of reinforcement learning (RL) in a multitask setting. Modern advances in deep RL are primarily associated with using model-free approaches operating in fully observable environments \cite{nature_dqn}. However, their practical use, for example, in mobile robotics \cite{Chaplot2020a,Staroverov2020b} and unmanned transport systems \cite{Yu2018}, is limited primarily due to sample inefficiency \cite{Zhu2020c}. This means that the agent needs to perform many actions in a complex, computationally expensive environment to generalize the incomplete information received and form a policy to achieve the goal. 
Improving task generalization is one of the central challenges in reinforcement learning \cite{maml,metaworld}, and moving in this direction will significantly expand the range of applications of modern RL methods.

Goal conditioned RL \cite{hindsight,OhSLK17}, Hierarchical RL \cite{hier_rl,Li2020a}, Meta-RL \cite{maml, rl2}, and model-based RL \cite{mbrl} should be considered as promising areas for solving the task generalization problem. It is the combination of the latter two approaches that are considered in this paper, with an emphasis on the tasks of training a robotic agent in a photorealistic environment.

The automatic learning and use of the agent-environment interaction model have shown its effectiveness in game \cite{mbrl,muzero} and robotic \cite{dreamer, mbpo} environments. Several options were proposed for using the environment model in the agent learning process: to generate additional ``virtual" agent steps in the environment \cite{dyna}, to simulate full episodes to more accurately approximate the Q-value \cite{alphazero,muzero}, and to directly learn the policy only on ``imaginary" trajectories \cite{planet, dreamer}. In all these cases, replacing the actual environment with the modeled one can significantly reduce the required number of interaction episodes.
However, replacing the original environment requires special attention to constructing compact and accurate models, so in model-based RL, the critical problem is developing a reliable way to represent and update the model. One of the most promising approaches is the use of neural world models \cite{WM}, in which recurrent architectures encode the dynamics and cause-and-effect relationships. The key task here, especially in the case of visual image-based states of the environment, is the correct formation of the latent space, in which temporal dependencies and transitions of the environment are modeled. Variational Autoencoders \cite{vae} have shown their effectiveness here.

In this work, we propose solving the task generalization problem and expanding the possibilities of using model-based RL, based on neural network world models, for the case of adapting experience from multiple tasks. 
Thus, we will deal with the case of forming a single agent policy in a sequence of different tasks. Each of these tasks would have its own goal. We seek to achieve faster learning of new tasks by reusing the knowledge gained while solving previous tasks. This is also highly related to the problem of sample efficiency; still, task generalization is a central scope of this work. We claim that the agent that is using prior tasks information (e.g. data, environment) can generalize better if its sample efficiency for the current task is higher compared to the agent that does not. As an example of such agents, we propose a new meta-learning algorithm working with model-based RL --- Retrospective Addressing for Multitask adaptation (RAMa). Our algorithm adapts the agent's accumulated experience in solving various tasks through a retrospective study of specially organized address memory. We propose formalizing the multitask adaptation problem in the form of a meta-level Markov decision process (meta-MDP). It formalizes a process for which we will form a separate policy for finding the necessary agent's behaviors from the current experience to solve a new task (see Fig. \ref{figM:2} with the general scheme of the algorithm).

In robotics learning, the key issue is the use of photorealistic environments with plausible physics of the robot's interaction with the environment. However, increasing the likelihood leads to a severe complication of the agent's state space and actions, complicating the problem statement for reinforcement learning. We show that the effective memory organization for world models allows us to increase the efficiency of training a model-based RL agent in such cases as well.

The work is organized as follows. After a brief overview of the related works in the field of world models, memory usage in RL, and adapting training samples, we give a formal problem statement for reinforcement learning with a recurrent state-space model (RSSM) \cite{planet}. Then, we go on to describe the basic algorithm of our method with an analysis of the principles of its operation. In the experimental part, we offer two series of primary experiments. The first experiment is devised to check the principal ability of the model to benefit from the use of prior data. In this experiment, we provide the model access to the experience of the same task and test it in the DMC suite \cite{dmcontrol}. In the second experiment, we show how the model can generalize across tasks to increase sample efficiency. We show the performance of the model in the Meta-World benchmark \cite{metaworld}, photorealistic NVIDIA Isaac environment, and DMC suite \cite{dmcontrol}.

The main results of the work can be summarized as follows:
\begin{enumerate}
    \item We adapt RSSM-like \cite{planet} world models for a multitask case to allow them to generalize across similar experiences from previously solved tasks within the domain to build a better world model and accelerate the convergence of an agent solving the current task.
    \item We propose an original addressing mechanism, whose training can be formalized in the form of a one-step meta-MDP.
    \item We show the actual effectiveness of the proposed method compared to using baseline model-based approaches without multitasking adaptation (such as PlaNet \cite{planet}, and Dreamer \cite{dreamer}) on DeepMind Control Suite and the Meta-World benchmark \cite{metaworld}.
    \item We demonstrate the possibility of using model-based RL with an addressing mechanism in a photorealistic robotic simulator, which opens up the opportunity of reducing the severity of the sample efficiency problem in the robotic domain.
\end{enumerate}

\section{Related Work}

A number of approaches leverage environment modeling, use the multitask approach for reinforcement learning, use memory to increase model capacity, or reweight training 
samples to yield better performance of the agent's learning process. 

\textbf{Multitask reinforcement learning.} Model-agnostic meta-learning \cite{maml} alleviates multitask reinforcement learning by training a model with meta-objective, optimizing weights to be easily fine-tuned toward any task. Compared to that, our approach solves the tasks sequentially. Plan2Explore \cite{p2e} is quite close to our work as it answers the same research question: \textbf{How to explore the environment so that the collected experience would be informative for future tasks?} The main difference is that P2E explicitly includes the exploration stage maximizing a latent disagreement. In contrast, we focus on how to save as much environment interaction as possible by reusing and adapting experiences targeted at particular tasks. RL$^2$ \cite{rl2} uses RNN to encode information about the task into a fixed-size latent vector. 

\textbf{Better models with memory.} Variational Memory Addressing Variational Autoencoder (VMA-VAE) \cite{vma}, introduced an additional learnable addressing latent variable into VAE \cite{vae} that was used to retrieve the data sample from memory in order to guide the generative process. Differentiable neural computers \cite{DNC} provide a read-write mechanism that allows explicitly memorizing, preserving, forgetting, and retrieving information from memory trained end-to-end. 

\textbf{Memory in reinforcement learning.} There is another branch of research similar to our work. It leverages memory for RL forming an \textit{episodic memory} \cite{episodic1,episodic2}. These models use continuously growing key-value storage where the key is a concatenated state-action pair representation, and the value is a Q-value estimate. The memory is used to build better action-value estimates using similarity search in the representation space. 
Compared to this approach, we store whole episodes into memory and set up memory to contain episodes that represent solutions for different tasks. MERLIN \cite{merlin} learns a world model with an auxiliary memory module that increases its predictive performance.

\textbf{Reweighting training samples.} The Data Valuation algorithm (DVRL) \cite{DVRL} is a meta-learning algorithm, which is close to our approach in the way that it trains the proposal distribution for data samples, optimized jointly with the main model. In contrast to our work, it uses one dataset to focus on robustifying the supervised model while we use multitask data built within one RL domain to focus on speeding up the RL algorithm on the current task. Learning-to-Reweight \cite{learning_to_reweight} is a model-agnostic approach to reweighting data samples that aim at increasing the robustness.

\section{Background}

\subsection{Reinforcement Learning}

We formulate the problem of reinforcement learning as the Partially-Observable Markov Decision Process (POMDP). Formally, POMDP is a tuple $\left<\mathcal{S}, \mathcal{A}, \mathcal{O}, P, R, O, \gamma\right>$,
where $\mathcal{S}$ is the set of states of POMDP, $\mathcal{A}$ is the set of actions, and $\mathcal{O}$ is the set of observations. 
The environment changes its state according to the conditional transition distribution $P(s'\mid s, a)$, but the agent only has 
access to the output of the observation function $o = O(s', a)$, $o\in \mathcal{O}$. The agent is defined by policy 
$\pi(a_t\mid o_{\leqslant t}, a_{<t})$. It interacts with a partially observable environment by taking actions 
on the environment and getting a next observation. We write $o_t, r_t \sim p(o_t, r_t\mid o_{<t}, a_{<t})$ as a shorthand for 
$s_t\sim P(s_t\mid s_{t-1}, a_{t-1}),\: o_t=O(s_t, a_{t-1}), \: r_t = R(s_t, a_{t-1})$. The goal of the agent is to 
maximize its expected discounted sum of rewards $\mathbb{E}_{\pi}\sum_t \gamma^t r_t$.

\subsection{Model-Based Reinforcement Learning with RSSM}

World models explicitly learn environment dynamics to generate novel experience \cite{dyna,WM}. Visual control tasks require efficient approaches to state prediction. When observations are non-Markovian, this can be achieved only by incorporating latent states. 
For latent dynamics learning, we use the RSSM \cite{planet}, which learns the dynamics by building a Markovian latent state for each timestep $s_t$ given previous action $a_{t-1}$ by autoencoding 
observations $o_t$ and rewards $r_t$,
which are non-Markovian. The world model consists of a representation model, 
or an encoder $q(s_t\mid s_{t-1}, a_{t-1}, o_t)$, a transition model 
$p(s_t\mid s_{t-1}, a_{t-1})$, an observation model $p(o_t\mid s_t)$, 
and a reward model $p(r_t\mid s_t)$. The representation and transition models share 
common parameters in the RSSM network \cite{planet}. The world model is trained 
to maximize the variational lower bound on the likelihood (ELBO) of the observed trajectory 
conditioned on actions  $\mathbb{E}_{p}\log p(o_{1:T}, r_{1:T}\mid a_{1:T})$. This is done 
by incorporating an approximate posterior model $q(s_t\mid s_{t-1}, a_{t-1}, o_t)$, which is 
known as a representation model. This model acts as a proposal distribution for states $s_t$. 

For behavioral training, we use the Dreamer agent \cite{planet, dreamer} which trains the policy and its value function by \textit{latent imagination}, 
an approach where  policy $\pi(a_t\mid s_t)$ is optimized by predicting both 
actions $a_t\sim \pi(a_t\mid s_t)$ and states $s_{t+1}\sim p(s_{t+1}\mid s_t, a_t)$ 
without environment interaction. In particular, it uses a state-value function 
estimate on the latent state $s_t$ on which the policy is 
trained to maximize, i.e. 
$\mathbb{E}_{\pi}\sum_t V_{\lambda}(s_t)$ 
where $V_{\lambda}(s)$ is a multi-step value 
estimate with a hyperparameter $\lambda$ which controls the bias variance trade-off \cite{Sutton1998}. As the transition and value 
models are parameterized by neural networks, we can backpropagate through the value function, the transition model, 
and the action sampling to compute symbolic gradients of the value estimates w.r.t. policy parameters.

\section{Multitask Adaptation by Retrospective Exploration}
\label{method_section}

Having a set of tasks $\mathcal{T}$, we train a model that proposes samples each solving $\tau\in \mathcal{T}$. Those samples are used to train the model-based agent. For each task in $\mathcal{T}$, the agent is initialized from scratch, then is trained using a proposal model, putting the gathered experience into the global multitask dataset $\mathcal{M}$. That is, all information about prior tasks is stored in $\mathcal{M}$ and in the parameters of the proposal model. In general, tasks may come from arbitrarily different environments. 
We, therefore, assume that environments for each of the tasks are semantically similar to each other and represent different aspects of the domain. Tasks may differ in 
their own task-specific reward function $R_{\tau}$ and also in the state transition 
distribution (e.g. for the robotic manipulation of objects with different shapes). 
In our case, we approach \textbf{the problem of multitask adaptation, which is the problem of adapting prior experiences that solve different tasks, to the current task}. We leverage a mechanism that retrieves experiences and uses them to accelerate the training of the reinforcement learning agent for the current task $\tau'\in T$. We emphasize that, in general, 
$R_{\tau} \neq R_{\tau'}$, where $\tau, \tau'$ are the tasks from $\mathcal{T}$ whose solutions are stored in $\mathcal{M}$. Therefore, we have chosen model-based RL as we may need to approximate $R_{\tau'}$ for a newly retrieved experience.

To enable the efficient selection of a multitask experience, we train the proposal distribution to score episodes from $\mathcal{M}$ and obtain the weights of categorical distribution to sample from. The samples are passed to the agent forming training batches. We refer to this model as the addressing model. For multitask adaptation, the addressing model is trained in a lifelong fashion. This is done by first training the MBRL agent with the addressing model on the first task $\tau_0\in \mathcal{T}$, with the same task experience being bootstrapped by the addressing model into the world model. Once the agent is converged, we proceed to the next task, with the agent being reinitialized, and the process repeats. The whole cycle is summarized in \autoref{alg1}.

\begin{algorithm}[h!]
\caption{Multitask Adaptation}
\SetAlgoLined
\label{alg1}
\SetKwInOut{KwIn}{Input}
\SetKwInOut{KwOut}{Initialize}

\KwIn{Set of tasks $\mathcal{T}$ with order $\texttt{ord}$}
\KwOut{Multitask buffer $\mathcal{M}$;\\
 Addressing model $f$;
}

\For{$\text{task}$ $\tau \in \text{ord}(\mathcal{T})$}{
    Initialize model-based agent M\;
    Train M to solve $\tau$ using $f$ with buffer $\mathcal{M}$; \tcp{Algorithm {\color{blue}{2}}}
}
\end{algorithm}

The adaptation is made by retrieving experiences relevant to the current task and similar to the current episode. To ensure the training episode will be relevant for the current task, we train the addressing model to select trajectories of experience which maximize the agent's performance on the current task. To select an episode that would be similar to one of the current tasks, the addressing model scores not just individual multitask trajectories but in comparison to the episode of the current task. \textbf{This forces the addressing mechanism to retrospectively explore the past experience seeking for promising behavior to accelerate the agent's training for the current task.}

We refer to our method as Retrospective Addressing for Multitask Adaptation (RAMa). As we build addressing to make retrospectively explored episodes promising for the current task, a successful strategy would be to select such an experience that maximizes the expected agent's performance over imagined rollouts. Therefore, starting from real states, the imagination would be conditioned on exploratory states as they were not collected to optimize the current task.


\subsection{Adressing Mechanism}

\begin{figure*}[t]
\centering
\includegraphics[width=0.7\linewidth]{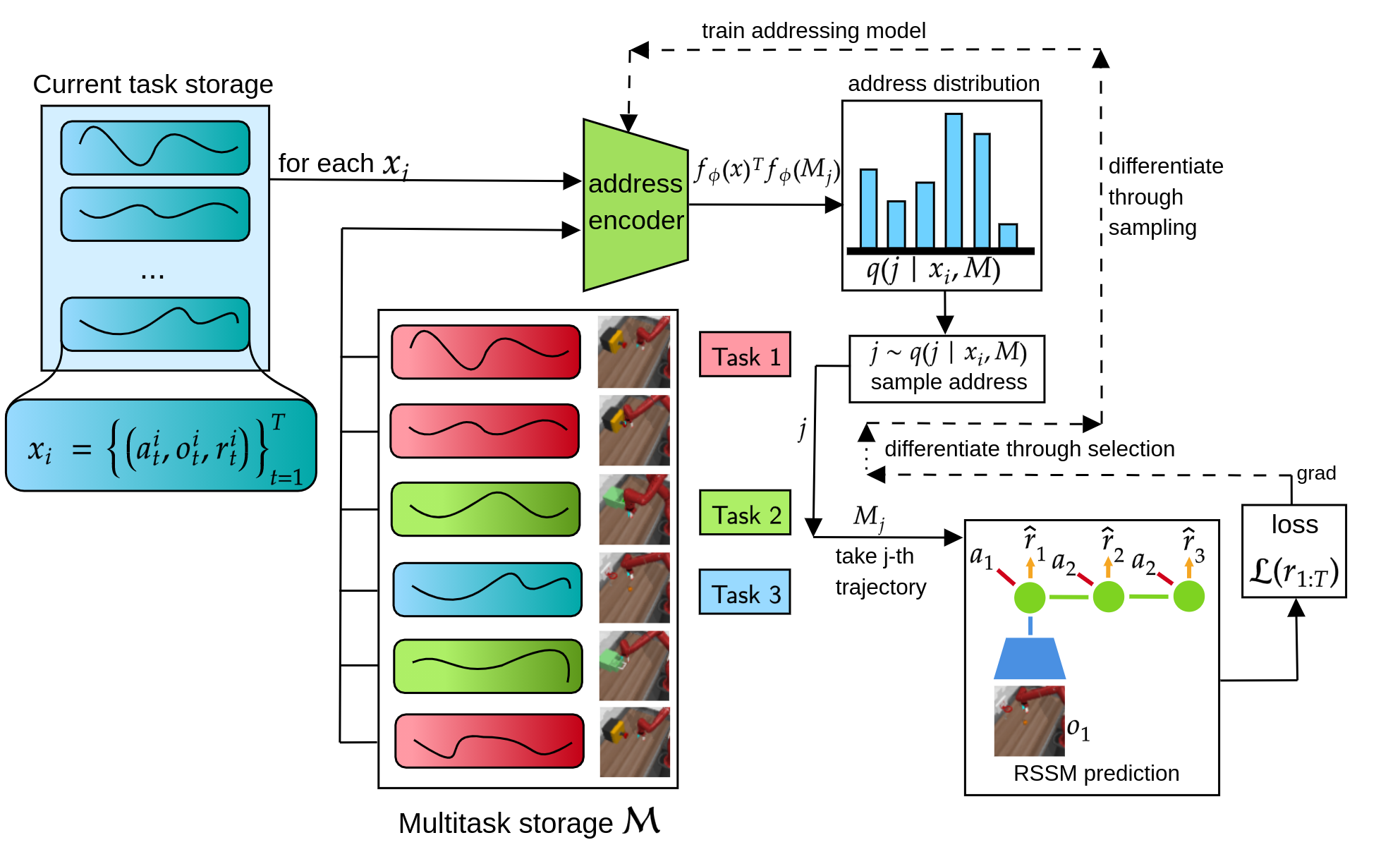}
\caption{A principal scheme for training Retrospective Addressing for Multitask Adaptation (RAMa).}
\label{figM:2}
\end{figure*}

For selecting multitask trajectories from $\mathcal{M}$, we use the learned addressing model parameterized by a neural network. The model compares trajectory $x = \{(a_t, o_t, r_t)\}_{t=1}^{T}$ from the current task with a batch of multitask trajectories $M$. The model projects $x$ and each $M_j$ into the embedding space and then calculates softmax logits for each $(x, M_j)$ pair as a dot product in the embedding space. To select the multitask trajectory, we sample index $j$ from distribution $q(j\mid x, M)$ that has the form:


\[
    q_{\phi}(j\mid x, M) \propto \exp(f_{\phi}(x)^Tf_{\phi}(M_j)),
\]
where $f_{\phi}$ is the learnable embedding of trajectory $x$ or $M_j$ parametrized 
by a recurrent neural network with parameters $\phi$. We implement $f_{\phi}$ as a recurrent neural network that consumes a sequence of actions concatenated with observation embeddings obtained from CNN. As the output, we use the last output vector of the recurrent network.
In other words, we project trajectory $x$ and each $M_j$ into the embedding space. Next, we calculate the inner product between the embedding of $x$ and the embedding of each $M_j$, and finally, we pass the resulting vector of the inner products to the softmax that gives probabilities for a categorical distribution. At preliminary experiments, we have observed that a trained CNN is crucial here as its presence significantly boosts the performance (see supplementary materials). This indicates that the features that are key to a good representation are not those that are required for good addressing.

\subsection{Expected Performance as an Objective for the Addressing Model}

We consider several approaches for training the address network. 
First, we feed each trajectory $M_j = \{(a^j_t, o^j_t, r^j_t)\}_{t=1}^{T}$, which is selected with respect to a particular trajectory $x_i$ (i.e. $j \sim q(j \mid x_i, M)$) to the RSSM and run latent imagination 
procedure to obtain its value estimates $V_{\lambda}(s^j_t)$. 
Then we sum the estimates over the imagination timesteps $t$ and backpropagate gradients of the resulting scalar loss up to 
the gradients w.r.t. the weights $\phi$ of the address network. We train the address network to minimize the objective:
\begin{gather}
    \mathcal{L}_{\text{V}} = -\mathbb{E}_{q(j\mid x_i, M)}\sum_{t}V_{\lambda}(s^{j}_{t})
    \label{V}
\end{gather}
To enable 
efficient gradient computation, we use the straight-through gradient estimator 
\cite{st_grad} to obtain differentiable samples $j$ returned as one-hot vectors. 
We then multiply each of these one-hot vectors by a batch of expert trajectories 
$M$ to obtain selection by index in a differentiable fashion. 
The overall process is outlined in Fig. \ref{figM:2}. We refer to this method as Value-based  Retrospective Addressing for Multitask Adaptation (VRAMa). 

The multitask batch $M$ is sampled from the multitask buffer $\mathcal{M}$. The current task batch $x$ is sampled from the same buffer sliced to only current task experience $\mathcal{M}_{\tau}$. The addressing model learns to select such trajectories from batch $M$, leading to high rewards according to the imagined experience. As the addressing model is trained to maximize value estimates over the selection, it will learn features specific to the reward function of the current task. And also, as the model projects both $x$ and $M$ into the embedding space using the same parameters $\theta$, it can adaptively perform selection, anchoring each $M_j$ around the current object $x_i$. These two properties combined lead to a model that clusters the trajectories in some space according to their behavior characteristics regarding the current task (e.g. by expected performance). This allows the model to select $M_j$  from the locality of $x$ in this space.

Another approach for training the address network is to use the REINFORCE algorithm 
\cite{reinforce}. For this, we sample $j \sim q(j \mid x, M)$. Then we predict 
the rewards of the selected trajectory $M_{j}$ using the Dreamer's reward model: 
$r_t^j\sim p(r_t\mid s^j_t)$, $s^j_t\sim q(s^j_t \mid s^j_{t-1}, a^j_{t-1}, o^j_t)$ and update the address network to minimize 
the objective: 
\begin{gather}
    \mathcal{L}_{\text{R}} = -\mathbb{E}_{q(j\mid x, M)}\left(\sum_t r^j_t\right) \log q(j \mid x, M)
    \label{R}
\end{gather}

We refer to this meta-objective as REINFORCE Retrospective Addressing for Multitask adaptation (RRAMa) with the RSSM. Note that, unlike the previous objective, this one is agnostic to the choice of the world model kind as it only requires the reward prediction. 

In this case, REINFORCE acts as a meta-objective for training a contextual bandit, which, in our case, is the addressing model. The bandit acts in the one-step meta-MDP where $x$ is a state object, $j$ is action, and reward is determined by $M_j$. To pose the objective into our framework of the expected performance maximization, a natural choice would be to meta-reward the bandit with the sum of predicted rewards along the trajectory $M_j$. This approach is also in line with our general paradigm of expected performance maximization by selection since $\sum_t r^j_t$ can serve as the estimate of the value of the whole trajectory for the current task.


Inspired by policy gradient algorithms \cite{Williams1992,Schulman2015}, we can subtract a state-dependent baseline from the bandit's reward. Since the baseline can depend only on state, a natural choice will be to use the same quantity but for $x$, i.e. the MBRL agent's estimate of episode $x$ reward. These rewards are sampled from the reward head $r_t\sim p(r_t\mid s_t)$, $s_t\sim q(s_t \mid s_{t-1}, a_{t-1}, o_t)$, where $x = \{(a_t, o_t, r_t)\}_{t=1}^{T}$. The baseline version of the previous objective:

\begin{gather}
    \mathcal{L}_{\text{RB}} = -\mathbb{E}_{q(j\mid x, M)}\left(\sum_t r^j_t - r_t\right) \log q(j \mid x, M)
    \label{RB}
\end{gather}

We refer to this meta-objective as REINFORCE baseline Retrospective Addressing for Multitask adaptation (RbRAMa) with RSSM. 

Another important observation about the approach is that we use the same model $f_{\phi}$ to encode both $x$ and each $M_j$. This means, for a fixed index $j$ the logit which is equal to $f_{\phi}(x)^T f_{\phi}(M_j)$ would not change if $x$ were used in place of $M_j$ and $M_j$ in place of $x$. However, the target which REINFORCE is training towards will change after such a swap. Despite the corresponding gradients w.r.t. $x$ and w.r.t $M_j$ still will differ from each other, this motivates us to use a detached version of $f_{\phi}(x)$ i.e. 
$\texttt{stop\_gradient}(f_{\phi}(x))$ in logit calculation. In this case, the addressing model will learn to adjust only the embedding of the $M_j$. 


\begin{table*}[t!]
\centering
\begin{tabular}{|c|c|c|c|c|c|c|c|c|}
\hline
& \multicolumn{3}{ |c| }{Walker} & \multicolumn{2}{ |c| }{Hopper} & Isaac \\
\cline{2-7}
& Stand & Walk & Run & Stand & Hop & Box \\
\hline
Dreamer & $896 $ & $802$ & $509 $ & $634 $ & $130$ & $12 $ \\
No addressing & $850 $ & $780 $ & $667 $ & $759 $ & $233 $ & $4 $ \\
RRAMa (Ours) & $902 $ & $\boldsymbol{869}$ & $\boldsymbol{695}$ & $737 $ & $\boldsymbol{238}$ & $\boldsymbol{15}$ \\
VRAMa (Ours) & $\boldsymbol{909}$ & $852 $ & $684 $ & $\boldsymbol{772 }$ & $176  $ & 14\\
\hline
\end{tabular}
\caption{In this experiment, buffer $\mathcal{M}$ was pre-filled with episodes obtained while solving the same task. For each task, we report the performance of the Dreamer baseline, VRAMa (\ref{V}), RRAMa (\ref{R}), and a variant of Algorithm \ref{alg2} without the addressing model (uniform sampling). For each run we calculated an average episode return over the course of training. Full return plots are shown in the Appendix.}
\label{same_task}
\end{table*}

\subsection{Training World Models with Address Network}

For training the MBRL agent using address network to accelerate the training, 
we modify the input batch as follows. Given the training batch 
$x = \{x_i\}_{i=1}^{n}$ and the multitask dataset $M = \{M_j\}_{j=1}^{m}$, 
for each $x_i$, we sample $\textbf{\textit{j}}_i \sim q(j\mid x_i, M)$ and feed the RSSM with an updated training batch $\{x_i\}_{i = 1}^{n}\cup\{M_{\textbf{j}_i}\}_{i=1}^{(1 - \beta)n}$, where $\beta$ is a fraction of the multitask experience. As the size of the multitask data may be tremendous, we can amortize the logits calculation by preserving the set of previously calculated addressing embeddings of $M$ for many training steps. Practically, we recalculate the whole set of multitask embeddings and use them for $N$ training steps and then the process repeats.
Once the batches are merged, we disable reward learning for parts of the batch that came from different tasks.  

We have found it crucial not just to score individual samples $M_j$ to obtain softmax weights but to do so in an adaptive way having $x$ in the condition of the addressing model. If the addressing model would just scored trajectories $M_j$ alone, it would converge to selection proportional to the current task reward of $M_j$. This is equivalent to just skew the selection of multitask trajectories toward higher rewards (obtained from the reward head). However, our experiments have shown that the most successful agent is not the one that gets the maximal predicted trajectory reward by selection. 

The whole process of training RSSM and address network
is described in Algorithm \ref{alg2} described in Appendix.


\section{Experimental Setup}
\label{exp_setup}


\textbf{DeepMind Control Environment.} We test our model on five visual tasks from DeepMind Control Suite \cite{dmcontrol} based on MuJoCo physics engine \cite{mujoco}. Observations of an Agent are $64\times 64\times 3$ images, the actions range from one to 12 dimensions, the rewards range from zero to one, the episodes last for 1,000 steps and have randomized initial states. The tasks include Walker Stand, Walker Walk, Walker Run, Hopper Stand, and Hopper Hop. These tasks represent different tasks within two domains, namely, Walker and Hopper. For each domain, we define a sequence of tasks with increasing difficulty. For the Walker domain, these are Stand, Walk, and Run. The multitask adaptation procedure will first solve Stand, then, with trajectories for the Stand task in the multitask buffer, solve the Walk task, and finally, having experience for the Stand and Walk tasks, solve Run. For the Hopper domain, the set of tasks consists of two tasks --- Stand and Hop.

\textbf{Metaworld} \cite{metaworld} is a set of robotic manipulation environments that offers both parametric and non-parametric task variability. After preliminary experiments, we adjusted camera positioning to make it more accessible for the visual agents. The agent observes  $64\times 64\times 3$ images, action is $4$-dimensional vector ranging in $[-1, 1]$. Meta-World offers a set of different manipulation tasks, e.g. picking, pushing, pulling objects and different parametrizations within each task (e.g. goal position). In the Meta-World benchmark, the main source of the difficulty is the intra-task variability, i.e. each episode had its goal or initial object positions slightly different, making the tasks much harder. In all experiments, we enabled such task variability, i.e. the environment samples 50 random vectors at startup and makes one of them active on each reset. Notably, between two different runs, the task vectors are different. 

\textbf{NVIDIA Issac Environment}. For our model, we built a custom continuous control environment implemented in NVIDIA Isaac SDK. The env has $96\times 64\times 3$ images as observations, $3$-dimensional continuous actions within the range $[-1, 1]$, and unbounded rewards, which can be both positive and negative. The env simulates a robot that has to move an object (e.g. a blue box or a blue cone) toward the goal position. The environment represents a simulated kitchen, with a robot standing near the table. The goal of the robot is to move a specific type of object to the goal zone, which is colored blue. One of the key features of this environment is the photorealistic rendering mode that uses a ray-tracing algorithm. We added the detailed description of the Isaac environment into the Appendix.



\textbf{Hyperparameters.} In all experiments, we implemented the address network with a GRU cell \cite{GRU}. We trained the address network with Adam \cite{adam} optimizer using a learning rate $0.001$. We left all non-address net specific hyperparameters to be default to the Dreamer algorithm \cite{dreamer}, i.e. we trained the model with batch size $n=50$; each batch consisted of sequences of length $L=50$. We trained the RSSM and the address net, each for $C=100$ optimizer steps between the episode collection. We set the $\lambda$-return parameter to be $0.95$ and $\gamma=0.99$. The probability of using multitask data in the training batch was $p=0.5$, the fraction of the multitask data in such batch was $\beta=0.5$. The number of the episodes for initial pre-training was $S=1$ (i.e. no pre-training on random episodes), and the initial size of the multitask buffer was set to $K=1000$ episodes (we randomly subsample the episodes resulted from the previous task training, only to decrease memory consumption). For the DeepMind Control Suite environments, we trained each model for 
$2000000$ environment steps. For Isaac Environment and Meta-World, each model was trained with $500000$ and $1000000$ environment steps, respectively. For all experiments within the Isaac domain, we increased imagination horizon from $h=15$ to $h=25$. We include the ablation study justifying hyperparameters in Appendix.

\section{Results}
\label{sec_results}

\begin{table*}[h!]
\centering
\begin{tabular}{ |c|c|c|c|c|c|c|c| }
\hline
& Push & Peg insert side & Sweep into &  Dial turn & Drawer close \footnotemark\\
\hline
Dreamer       & $301\pm 185$ & $206 \pm 129$ & $441\pm 353$ & $2805 \pm 575$ & ${4750 \pm 70}$ \\
No addressing & $243\pm 186$ & $\boldsymbol{468 \pm 290}$ & $247\pm 152$  & $2117 \pm 852$ & ${4743 \pm 85}$ \\
RRAMa (Ours)  & $\boldsymbol{443\pm 301}$ & $38 \pm 26$ & $338\pm 157$ & $\boldsymbol{3051 \pm 464}$ & ${4622 \pm 0.1}$ \\
VRAMa (Ours)  & $121\pm 82$ & $299\pm 245$ & $266\pm 74$ & $2938 \pm 365$ & ${4467 \pm 0.1}$\\
\hline
\end{tabular}

\caption{In the simple multitask experiment for the Meta-World domain, we first collected a number of episodes, each corresponding to one of 50 different vector task parameters. Then, we launched the baselines and our proposed methods with completely new vector task parameters. For Dial turn and Push domains, the addressing model was able to generalize best compared to the baselines. The full return curves are shown in the Appendix.}
\label{multitask_metaw}
\end{table*}

\begin{table*}[t!]
\centering
\begin{tabular}{ |c|c|c|c|c|c|c|c| }
\hline
& \multicolumn{2}{ |c| }{Walker} & Hopper & NVIDIA Isaac \\
\cline{2-5}
& Stand $\rightarrow$ Walk & Walk $\rightarrow$ Run & Stand $\rightarrow$ Hop & Box $\rightarrow$ Cone \\
\hline
Dreamer & $786\pm 112$ & $509 \pm 34$ & $130\pm 63$ & $4\pm 0.01$ \\
RRAMa (Ours) & $823\pm 30$ & $\boldsymbol{618 \pm 24}$ & $174\pm 45$ & $\boldsymbol{9 \pm 0.01}$ \\
VRAMa (Ours) & $\boldsymbol{843\pm 20}$ & $550\pm 23$ & $\boldsymbol{182\pm 11}$ & --\\
\hline
\end{tabular}

\caption{For the multitask experiment, we filled the $\mathcal{M}$ with an easier task experience and trained the agent to solve a harder one. For each run we calculated an average episode return over the course of training. Full return plots are shown in the Appendix.}
\label{multitask_t}
\end{table*}

\subsection{Same Task Experiment}
\label{st_exp}

To\footnotetext{For Drawer close task, all models were able to solve the task with equal performance. For Sweep into, none of them have solved the task.} test the ability of our model to reuse prior task information in principal, we tested the addressing model the case when prior episodes solve the same task as does the current agent. For the Walker domain, we ran each of three tasks: Stand, Walk, Run first using vanilla Dreamer. After training Dreamer we initialized $\mathcal{M}$ with $1000$ episodes obtained while training. Also, we filtered $\mathcal{M}$ to have only episodes with high episode returns. For the walk task, the filter threshold was set to $600$ for the stand task in both Walker and Hopper domains it was set to $0$, i.e. no filtering. For the Isaac domain, we also did not filter prior data. Using this buffer, in each task, we launched VRAMa, RRAMa, and a vanilla Dreamer but with buffer prefilled with $\mathcal{M}$. We repeated each run twice and reported the average train return with standard errors. For the Hopper Hop task, we repeated the runs four times as it has much higher variance. The results are shown in Table \ref{same_task}. For each domain and algorithm, we report the average episode return over the course of training. For this and the following experiment, each run took around 24 hours on Titan RTX GPU and 10 CPUs.

Though the main improvement against the baseline was made by just giving the world model access to high-quality trajectories, our approaches constantly show improvement over two baselines for Hopper and Walker domains. Although this experiment should be considered as a toy experiment, it indicates that the model can still gain performance by just reweighting the training distribution. Also, the variance for our methods is lower than for baselines, indicating that they are less vulnerable to stochasticity. In contrast, for the Isaac domain, RRAMa outperformed both baselines significantly. Importantly, the no addressing baseline performed no better than vanilla Dreamer, which means, for this domain, it is not enough just to provide the agent with higher quality episodes, and there is a necessary stage of adaptive batch sampling.

\subsection{Multitask Experiment}

To see how the proposed model would generalize across the tasks, we first trained the agent to solve simpler tasks. We performed multitask experiments on the DMC, Isaac, and Meta-World benchmarks. For the Hopper domain, we collected experience of the Stand task and then used it as multitask buffer for the Hop task. As in the previous experiment, we trained VRAMa, RRAMa, and two baselines. We averaged each model performance over two runs. Both of our approaches outperformed the baseline, which indicates the positive performance of the addressing model.

For the Meta-World \cite{metaworld} domain, we tested the ability of the model to generalize over a parametric task variability. That is, we fixed the general goal (such as pushing or turning) and changed only the task parameter vector encoding starting and goal object position. Importantly, we did not provide the task parameter vector to the model explicitly. The addressing was able to gain the performance over the vanilla Dreamer for the Push, Dial-turn, and Peg-insert-side tasks. For Push and Dial-turn, the addressing outperformed two baselines. We note that the addressing is able to perform better for the tasks of the intermediate difficulty. For the Sweep-into domain, we do not mark any algorithm in bold as all of them failed on this task due to its complexity. On the other hand, the Drawer-close task is too simple to get any meaningful speedup. Therefore, all algorithms are in bold. The results are shown in Table \ref{multitask_metaw}. Averaged train episode returns are shows. Full plots are presented in the Appendix.

For the Isaac domain, we collected experience of the Bluebox task and then changed the shape of the target object from blue box to blue cone. The Bluebox task experience was used as the multitask storage in this case. For the Bluecone task, we ran the RRAMa and the vanilla Dreamer models. We additionally turned off the world model training for addressed batches, which we found beneficial for this domain. The results are presented in Table \ref{multitask_t}.

\subsection{Discussion}
\label{sec_discussion}

The obtained results suggest that the addressing model can indeed generalize among different tasks. However, a successful generalization requires overcoming the phenomenon known as the \textit{catastrophic forgetting} \cite{cat_forg}. Therefore, one should carefully apply the suggested method in the case of a higher number of tasks. We presented results on the novel NVIDIA Isaac environment and recent Meta-World benchmark, which we found much harder to solve than the DeepMind Control Suite. We plan to investigate the addressing model for these robotic domains to build a more robust addressing model.  

\section{Conclusion}

In this paper, we considered the problem of generalization between the tasks when solving a sequence of different tasks. We have proposed a new model-based reinforcement learning RAMa method that uses Retrospective Addressing for Multitask Adaptation. RAMa adapts the accumulated experience in solving various tasks through a retrospective study of specially organized address memory. We propose formalizing the multitask adaptation problem in the form of a meta-MDP, for which the agent forms a separate policy for finding the necessary agent's behaviors from the current experience to solve a new task. We have suggested two ways to train the mechanism for selecting previous cases from address memory: based on value (VRAMa) and REINFORCE (RRAMa). The proposed approach is one of the first attempts to use a trainable mechanism for selecting behavior cases using recurrent neural world models. Using the DMC suite, we demonstrate the effectiveness of using information from solving previous problems to speed up the construction of a policy to achieve a new goal. We have also demonstrated the efficiency of the RAMa method in robotic manipulation environments.

\bibliography{full}

\newpage 

\appendix

\section{NVIDIA Isacc Environment}
\label{appA}

In this work, we created a novel environment based on the NVIDIA Isaac Platform\footnote{https://developer.nvidia.com/isaac-sdk}. This platform allows many scenarios for the robotic AI research through the photorealistic simulation. The simulator runs real-time ray tracing to enhance the visual quality of the simulation, thus allowing the better AI models.

\begin{figure}[h]
\centering
\includegraphics[width=0.7\linewidth]{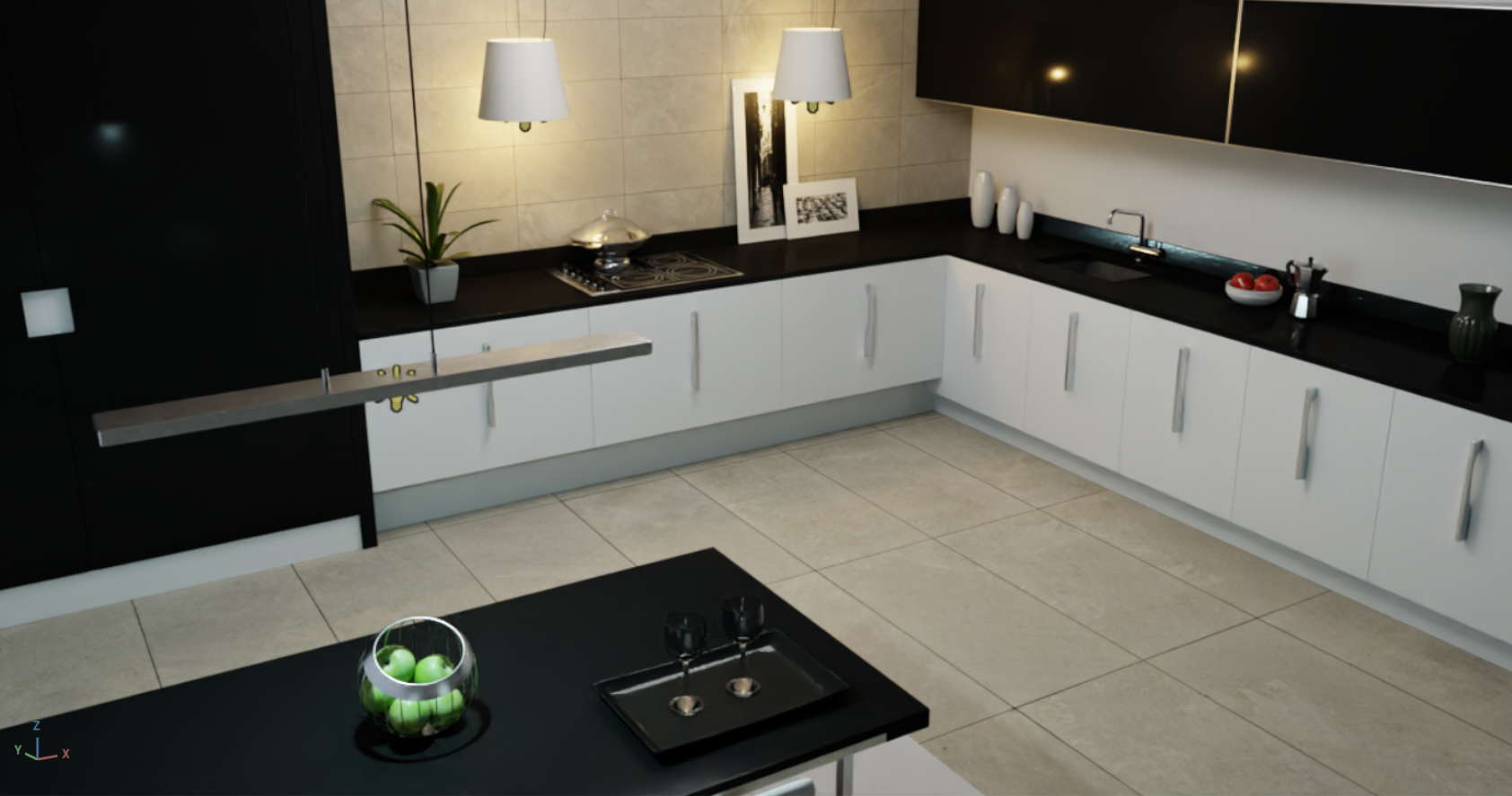}
\caption{Simulated kitchen from NVIDIA Isaac Simulator}
\label{kitchen}
\end{figure}

\begin{figure}[h]
\centering
\includegraphics[width=0.48\linewidth]{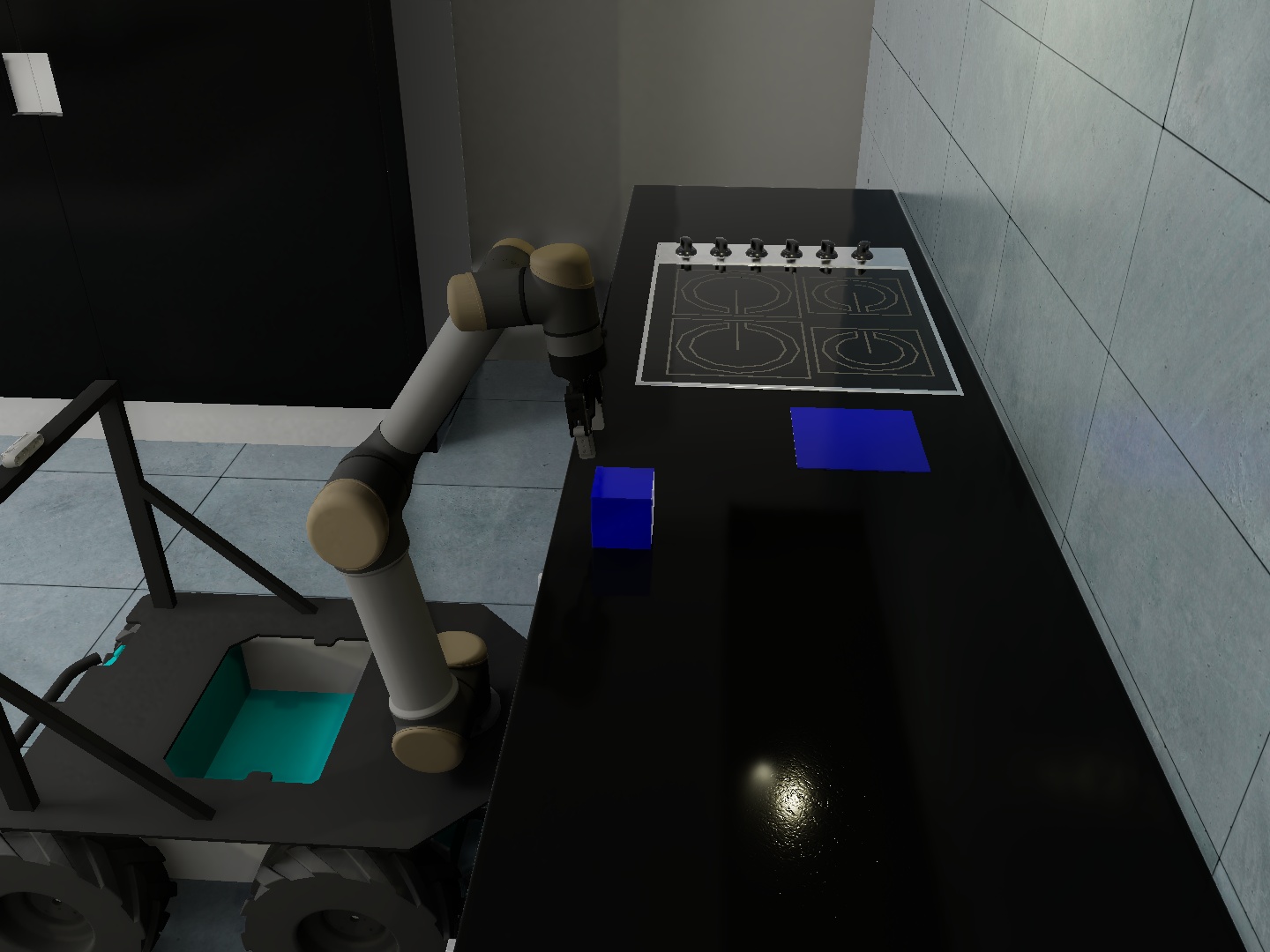}
\includegraphics[width=0.48\linewidth]{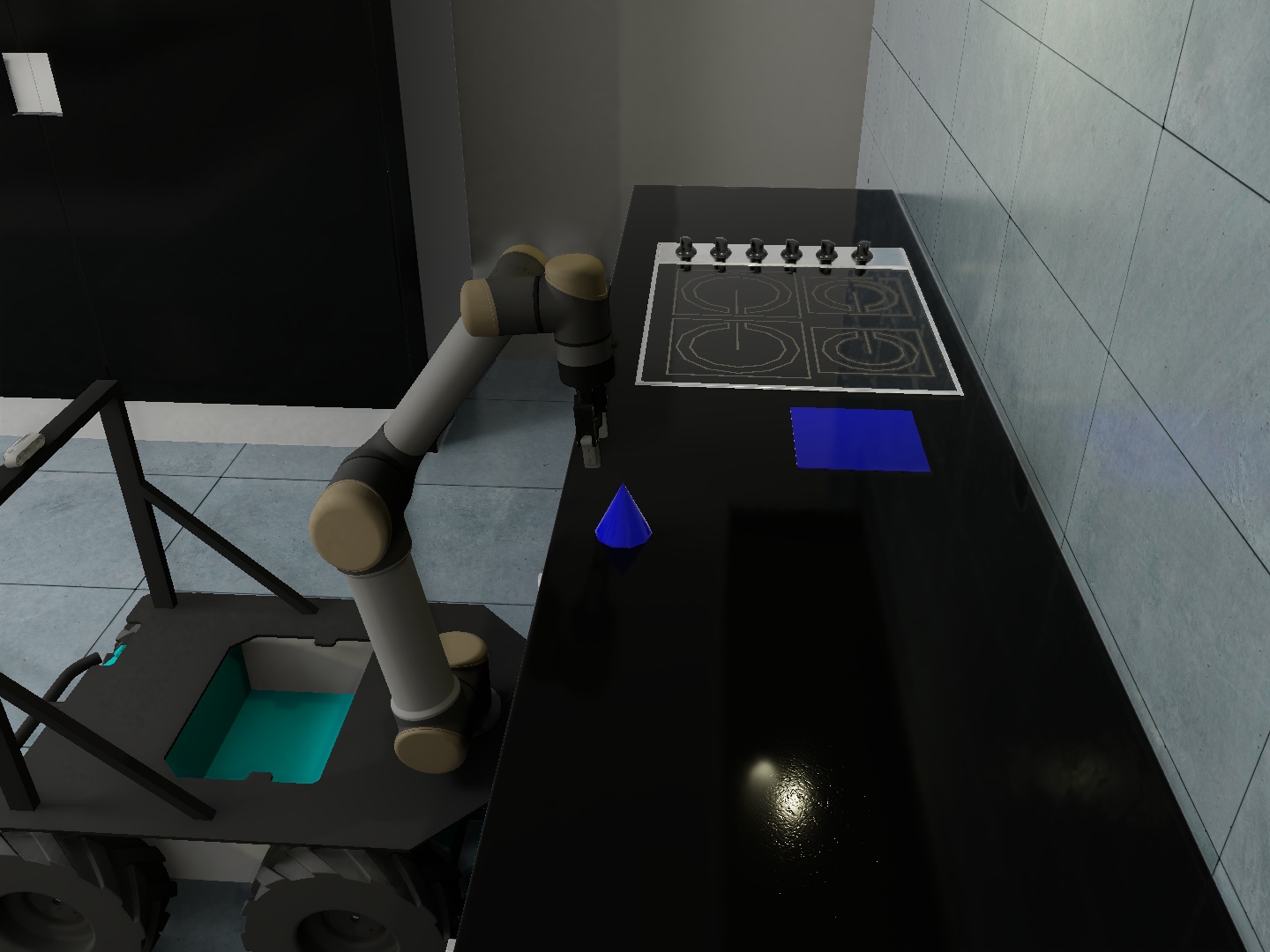}
\centering
\caption{Examples of the robot acting in our environment. Images are rendered at higher resolution than the agent observes. The left observation represents blue box target and the right observation represents blue cone target.}
\label{obs_ex}
\end{figure}

The environment is a simulated kitchen depicted in Figure \ref{kitchen}. The kitchen is one of the standard public assets of NVIDIA Isaac Simulator. Inside a simulated room there is a Husky robot with one UR5 arm. 
The robot has 7 joints, namely a pan joint which lifts the whole arm and down, 
rotation joint which can rotate arm around vertical axis, elbow joint which 
rotates middle half of an arm, 
three knuckle joints each corresponding to each rotational axis of the arm. 
The last joint is finger joint controlling object grasping of robot arm. 
The agent can output 7-dimensional vector with each component being inside 
range $[-1, 1]$ which is an acceleration at a particular 
timestep of a particular arm's joint. The observation of an agent is an 
RGB image of size $96\times64$ where camera is pointing at robot which stands 
before the table in a simulated kitchen. We parameterize our environment by the goal which agent pursues.
In the simplest case, the agent have small blue box on the table and it has to move this box on the target zone which is represented as blue rectangle area over table. In Figure \ref{obs_ex} there are two examples of observations for two different goals in our env rendered at higher resolution. In the first goal, agent needs to move blue box and in the second one, agent needs to move blue cone towards the target. Each step the agent receives reward equal to the difference between current distance between the box and the target position and such distance in the previous step. The maximal reward that the agent can obtain is equal to 50 which is the distance between the initial object position and the center of the target zone. We set the maximum number of steps for episode equal to 500. During initial experiments we found full 7-dims action space to be unstable for this task. Therefore, we disabled all joints except pan joint, lift joint and elbow joint. In this case, the object can be just pushed towards the zone by robot's knuckle.

\section{Addressing algorithm}

Algorithm \ref{alg2} shows the learning workflow of the addressing model that learns towards a specific model-based agent. The agent trains by extracting short contiguous ''chuncks`` of episodes of length $L$ and with batch size $n$. The learning process is divided into two stages:  training the addressing model given a fixed Dreamer model and training the Dreamer with fixed addressing model. In the first stage, we sample a batch of current task experiences $\{x_i\}_{i=0}^{n}$ and a batch of multitask experiences $\{M_j\}_{j=0}^{m}$. Then we use the addressing to adaptively score  each $M_j \mid x_i$. We sample from each of the corresponding distributions and calculate the loss of the addressing model according to VRAMa, RRAMa, or RbRAMa objectives. Finally, we backpropagate the loss to update the weights of the addressing model. In the second stage, we first recalculate the set of addressing latents for each chunk of length $L$ from the multitask dataset. Once they are calculated, we take the current task batch and calculate the addressing probabilities $q(j \mid x_i, M) \propto f_{\phi}(x)^T D_j$ where $D_j = f_{\phi}(M_j)$ -- the matrix of the addressing latents. To accelerate the algorithm, we don't recalculate the matrix of the addressing latents at each step, instead we do so only once per $C$ steps.

\begin{algorithm*}[h!]
\caption{Retrospective Addressing for Multitask adaptation (RAMa) with RSSM}
\label{alg2}
\SetKwInOut{KwIn}{Input}

\SetAlgoLined
\begin{tabular*}{\textwidth}{l @{\extracolsep{\fill}} p{\textwidth}}
\begin{minipage}{.7\textwidth}
\KwIn{Multi-task buffer $\mathcal{M}$ with $K$ episodes representing solutions for prior tasks from $\mathcal{T}$;}
  \While{not converged}{
  \For{train step $c = 1..C$}{
    \tcp{\texttt{address network training}}
    Draw $n$ current task trajectories $x_i \sim \mathcal{M}_{\tau}, \:i=\overline{1,n}$\;
    Draw $m$ multitask trajectories $M_j \sim \mathcal{M}, \:j=\overline{1,m}$\;
    \If{$c = 1$}{
        \tcp{\texttt{amortize large multitask memory calculation}}
        Store embeddings into matrix $D:$  $D_j = f_{\phi}(M_j)$, for each $M_j \sim \mathcal{M}$\;
    }
    Compute samples from address distribution for each $x_i$: $\textbf{\textit{j}} \sim q(j \mid x_i, M)$\;
    Compute reward, value estimates $r_{\textbf{\textit{j}}, t}$, $V_{\lambda}(s_{\textbf{\textit{j}}, t})$ using selected expert trajectories $M_{\textbf{\textit{j}}}$\;
    Backprop value from Eqs. \ref{V}, \ref{R} or \ref{RB} and update address net parameters $\phi$.\;
    \tcp{\texttt{World model training with addressing}}
    Set input batch $B\leftarrow \{x_i\}_{i = 1}^{n}$\;
    Sample new indices $\mathbf{j} \sim \hat{q}(j \mid x_i, M) \propto f_{\phi}(x)^T D_j$, select new $M_{\textbf{j}}$\;
    $\alpha\sim \text{Bernoulli}(p)$\;
    \If{$\alpha = 1$}{
     Set input batch $B\leftarrow \{x_i\}_{i = 1}^{\beta n}\cup\{M_{\textbf{j}_i}\}_{i=1}^{(1 - \beta)n}$\;
    }
    Update parameters $\theta$ of RSSM and the policy using batch $B$\;
  }
  Collect episode using the RSSM and the policy and store it to $\mathcal{M}$\;
  }
\end{minipage}
& \vskip -178pt
\begin{tabular*}{0.28\textwidth}{l @{\extracolsep{\fill}} l}
\textbf{Hyper parameters}\\ 
Sequence length & $L$ \\
Collect interval & $C$ \\
Batch size & $n$ \\
Multitask batch size& $m$ \\
Initial episodes & $K$ \\
Multitask batch proba & $p$ \\ 
Multitask fill fraction & $\beta$  \\
$\lambda$-return parameter & $\lambda$ \\
Current task & $\tau$ \\
Set of prior tasks & $\mathcal{T}$ \\
\end{tabular*}
\end{tabular*}
\label{alg2}
\end{algorithm*}

\section{Reward Curves}
\label{appCurves}

\begin{figure*}[ht]
\captionsetup[subfigure]{labelformat=empty}
\makebox[\textwidth][c]{
\subfloat[]{
  \includegraphics[width=0.35\linewidth]{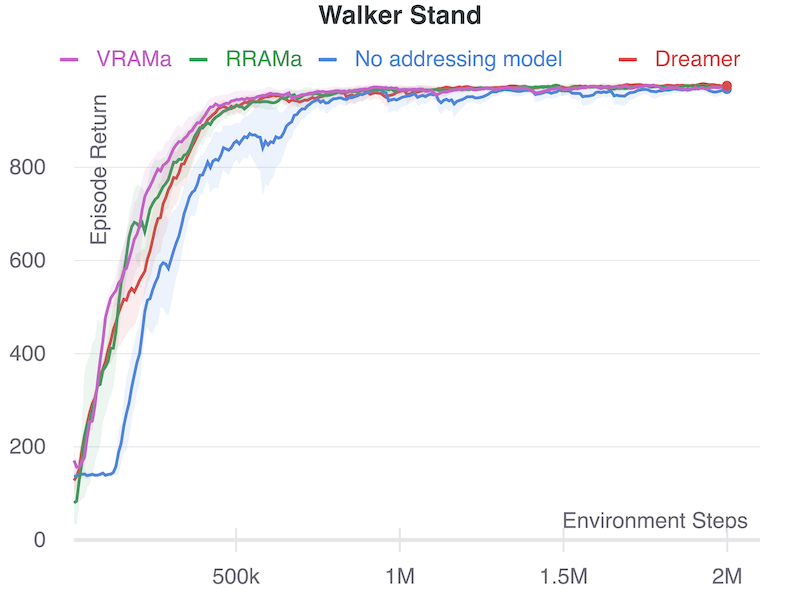}
  \label{wstand}
}
\hspace{-5mm}
\subfloat[]{
  \includegraphics[width=0.35\linewidth]{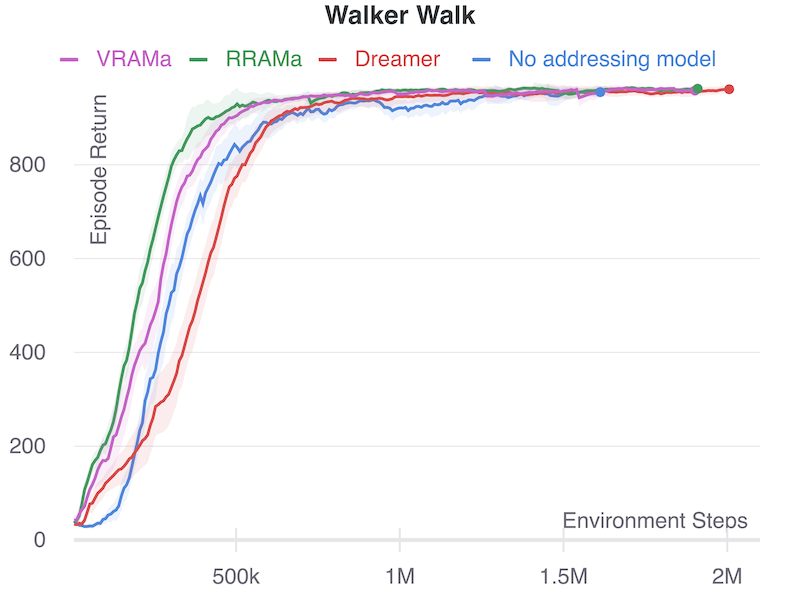}
}
\hspace{-5mm}
\subfloat[]{
  \includegraphics[width=0.35\linewidth]{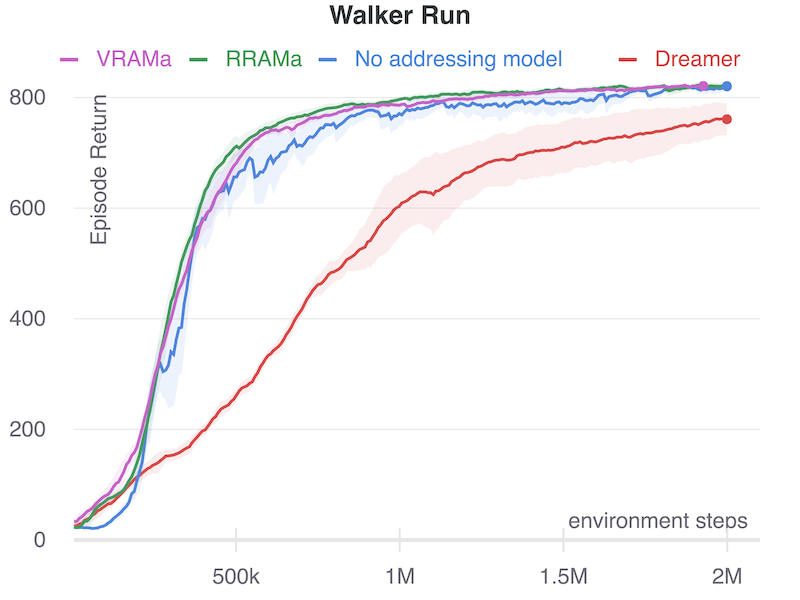}
}}
\vspace{-10mm}

\makebox[\textwidth][c]{
\subfloat[]{
  \includegraphics[width=0.35\linewidth]{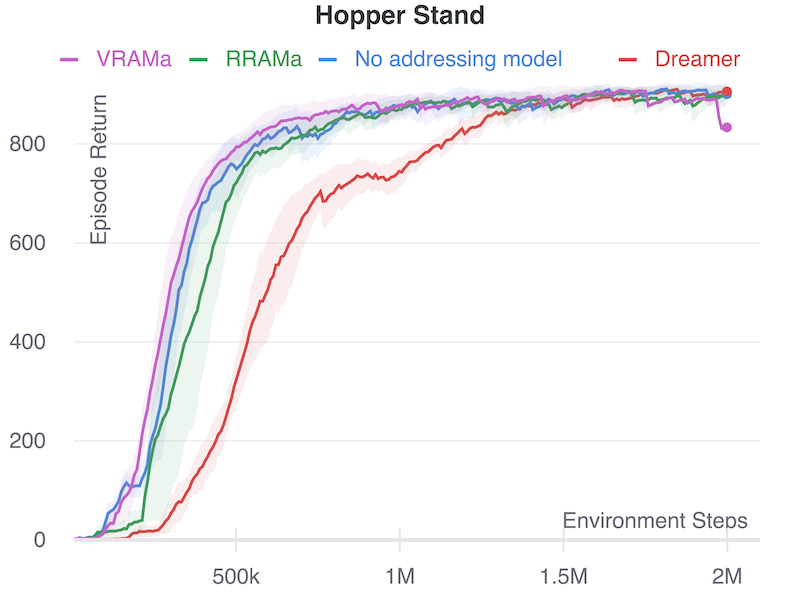}
}
\hspace{-5mm}
\subfloat[]{   
  \includegraphics[width=0.35\linewidth]{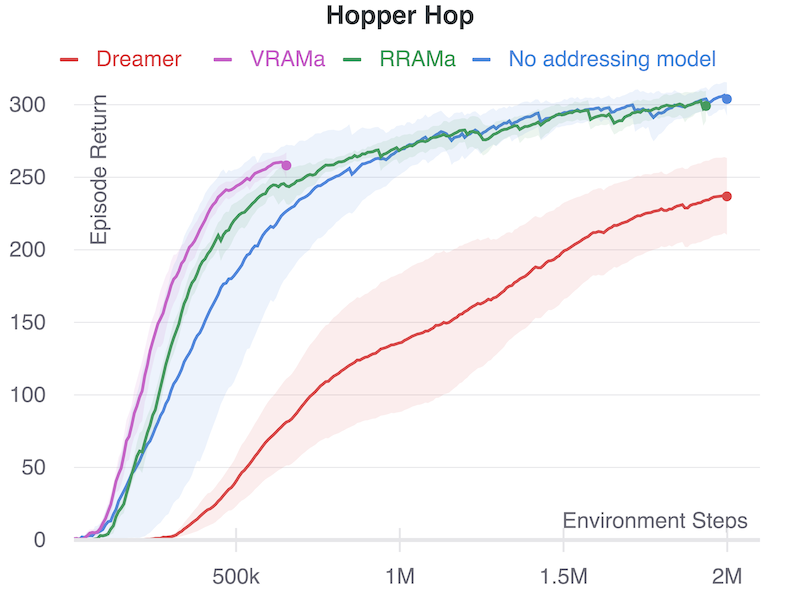}
}
\hspace{-5mm}
\subfloat[]{   
  \includegraphics[width=0.35\linewidth]{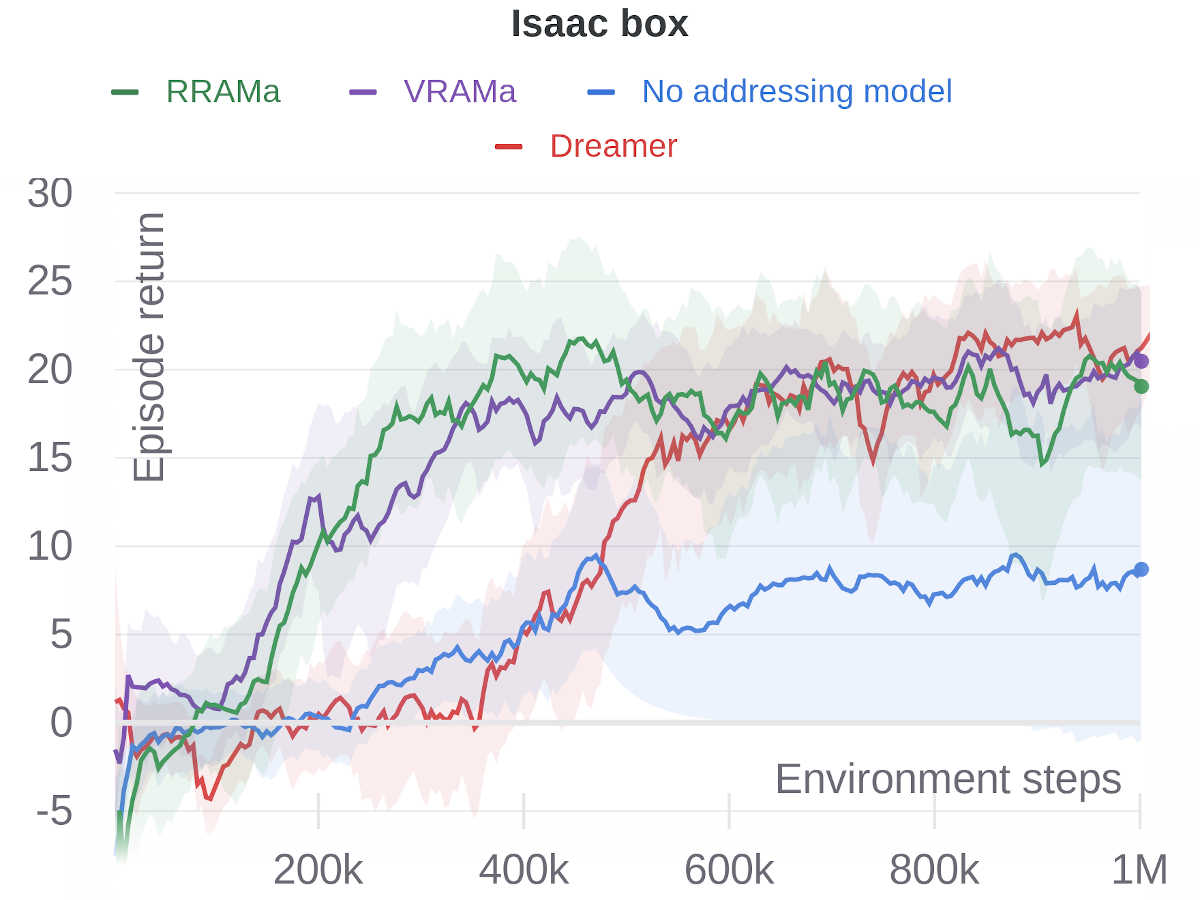}
}
}
\caption{Performance plots for same task experiment. For each task, we report the performance of the Dreamer baseline, VRAMa (\ref{V}), RRAMa (\ref{R}), and a variant of Algorithm \ref{alg2} without the addressing model (uniform sampling).}
\label{same_task_f}
\end{figure*}

\begin{figure*}[h]
\captionsetup[subfigure]{labelformat=empty}
\makebox[\textwidth][c]{
\subfloat[]{
  \includegraphics[width=0.35\linewidth]{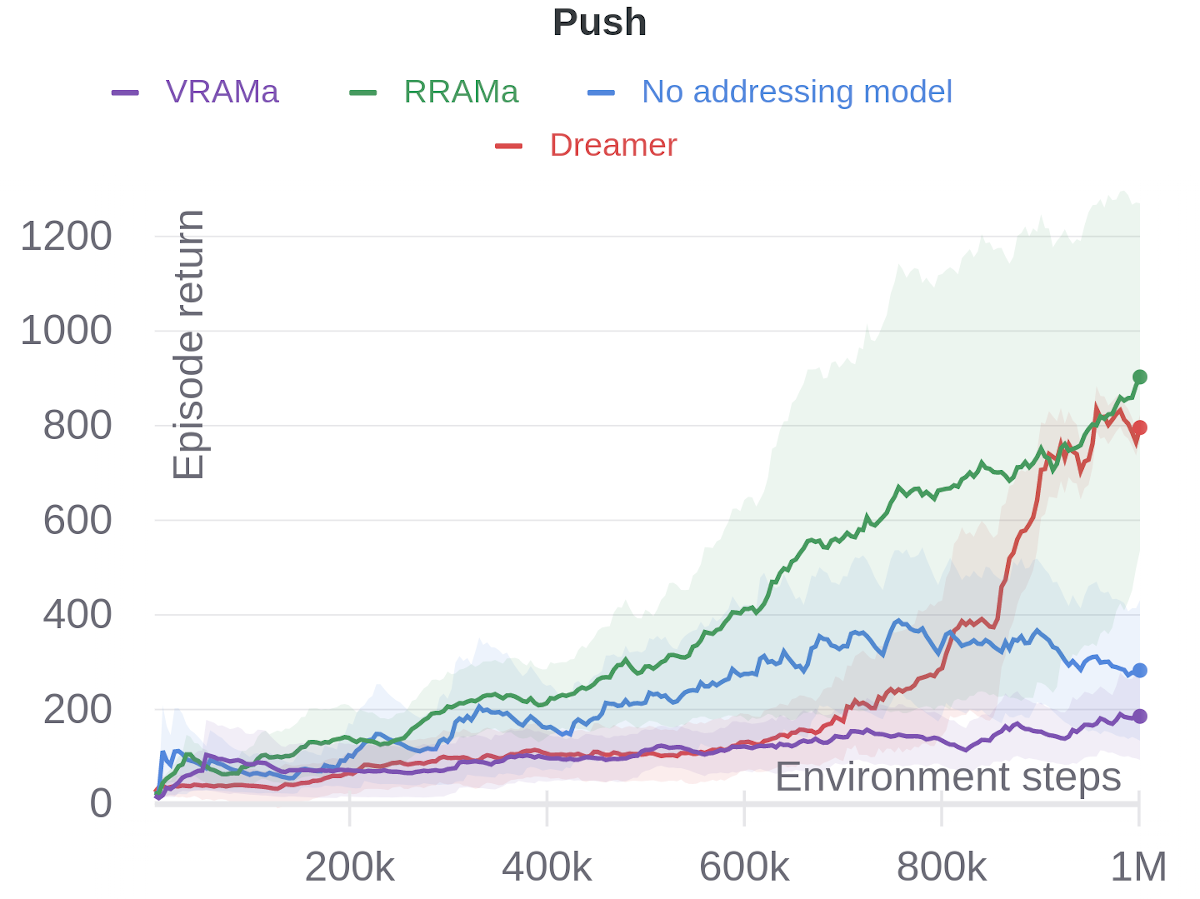}
  \label{push}
}
\subfloat[]{
  \includegraphics[width=0.35\linewidth]{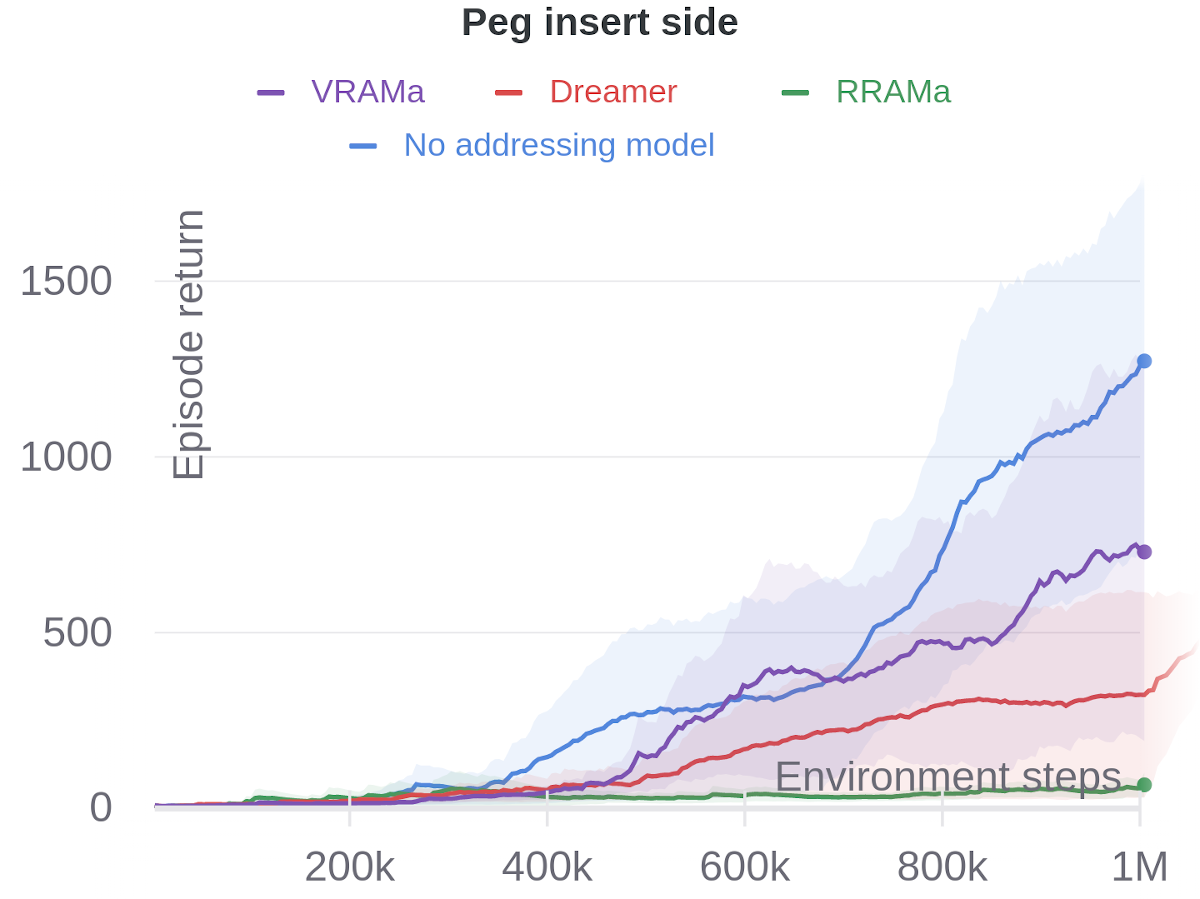}
}
\subfloat[]{
  \includegraphics[width=0.35\linewidth]{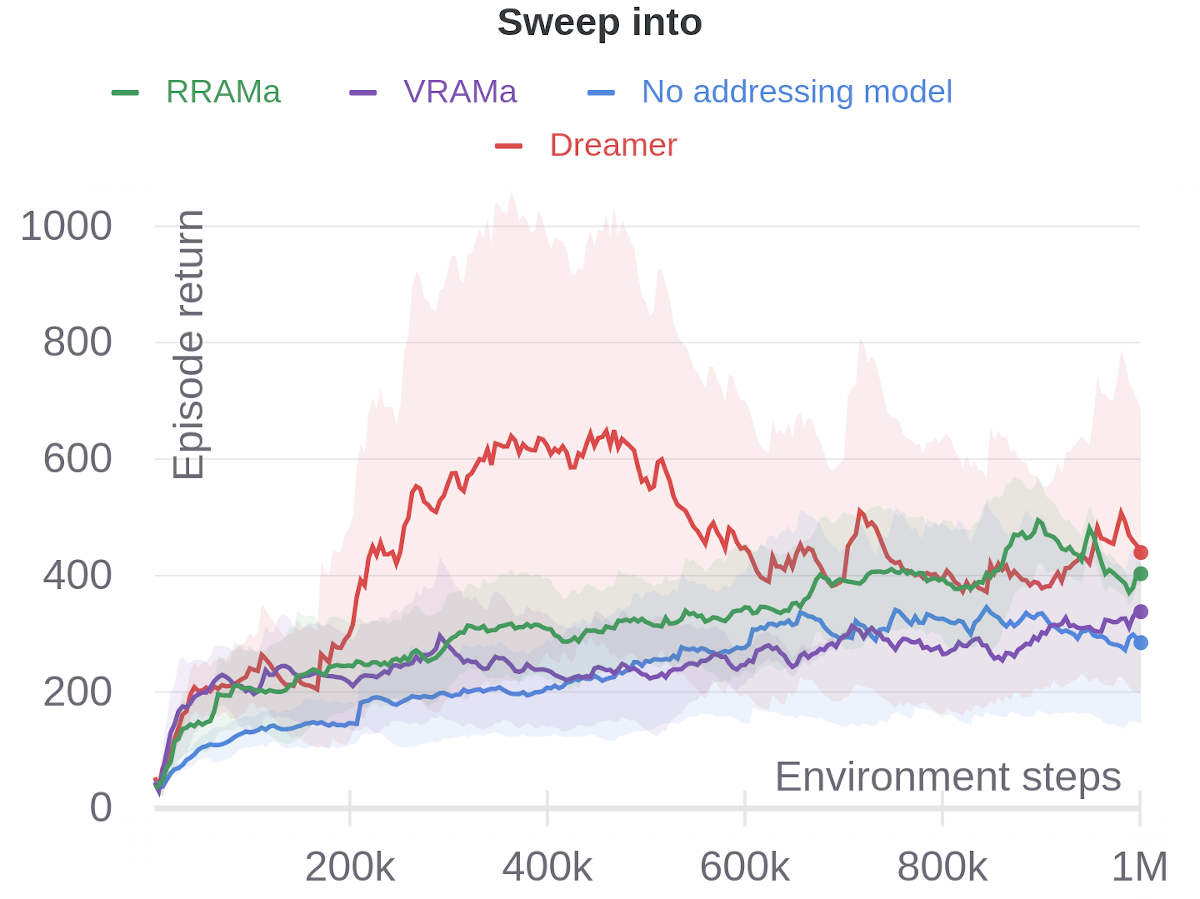}
}}
\vspace{-5mm}

\makebox[\textwidth][c]{
\subfloat[]{
  \includegraphics[width=0.35\linewidth]{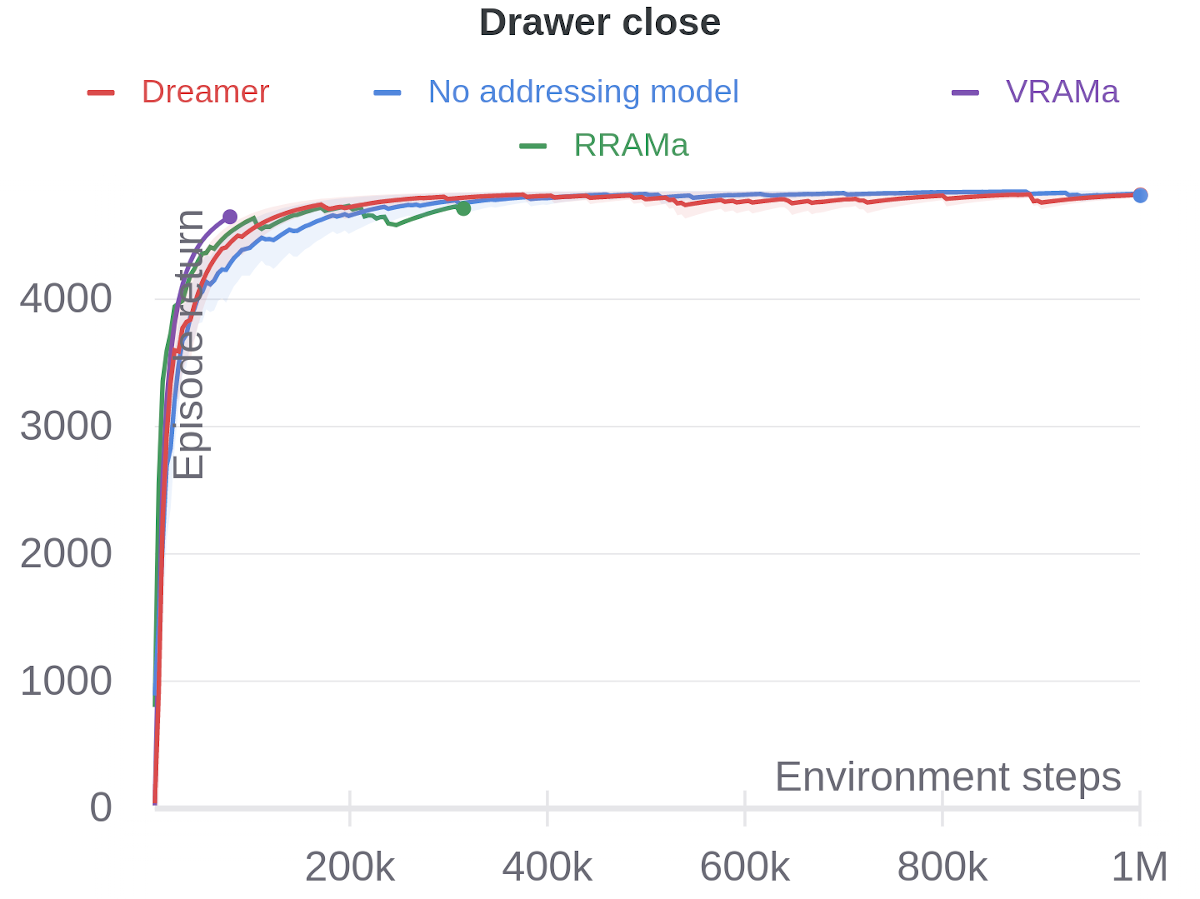}
}
\subfloat[]{   
  \includegraphics[width=0.35\linewidth]{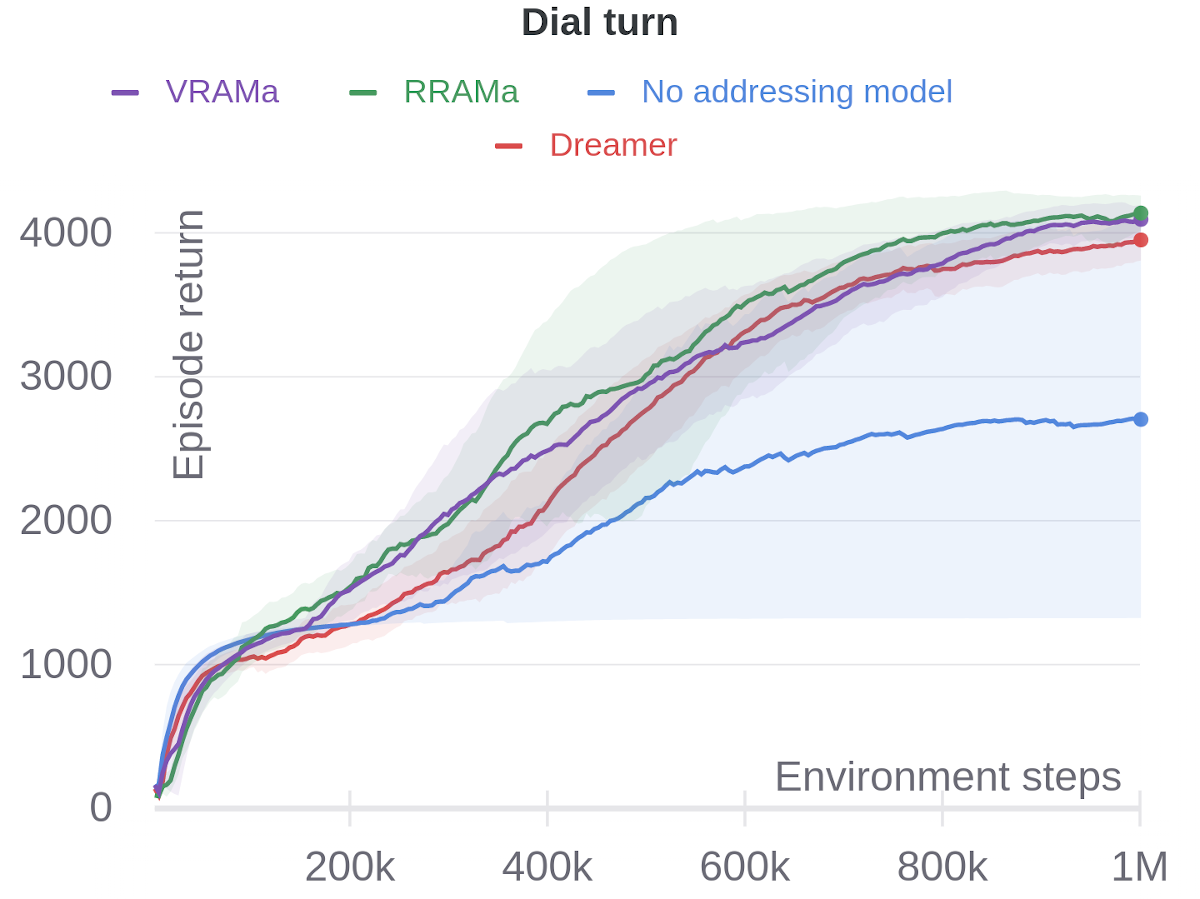}
}
\subfloat[]{   
  \includegraphics[width=0.35\linewidth]{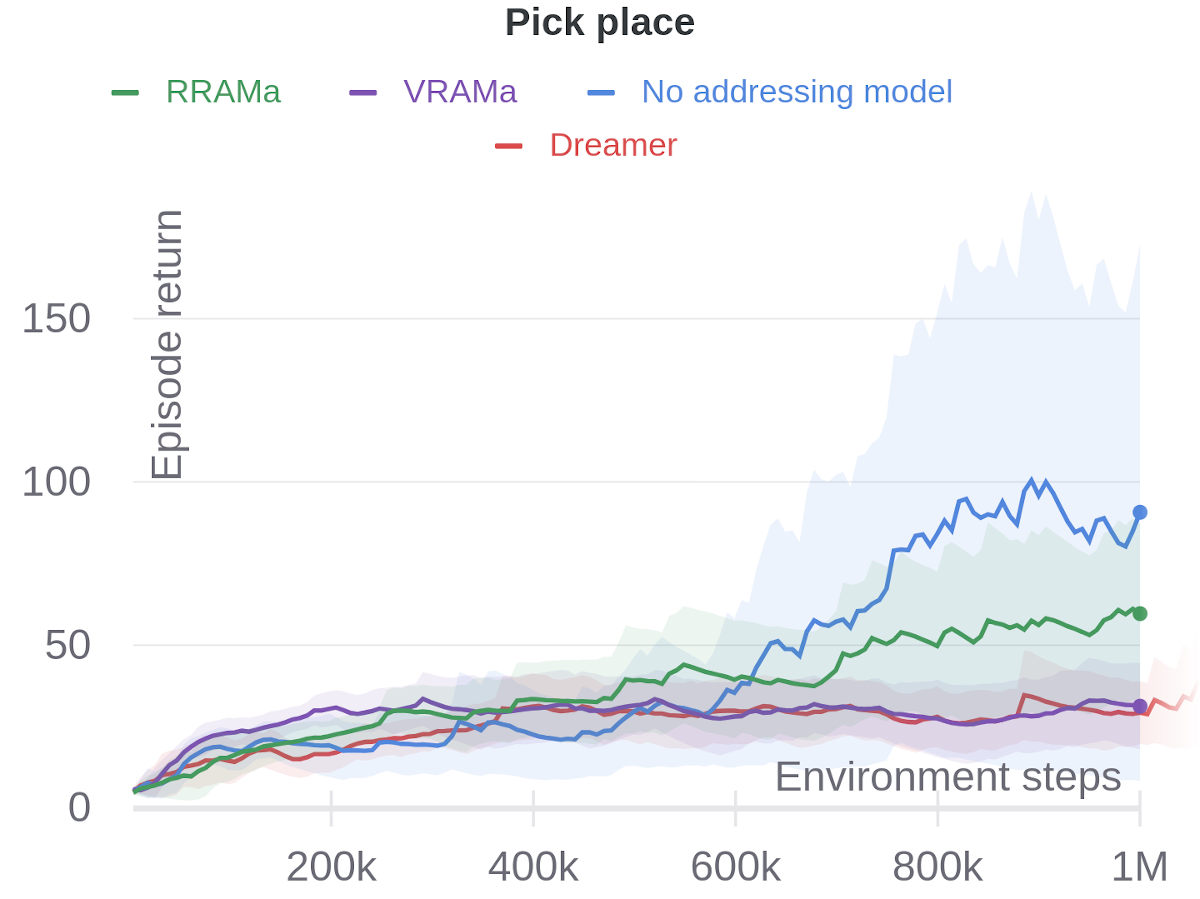}
}
}
\caption{Performance plots for same task experiment in Metaworld domains. For each task, we report the performance of the Dreamer baseline, VRAMa (\ref{V}), RRAMa (\ref{R}), and a variant of Algorithm \ref{alg2} without the addressing model (uniform sampling).}
\label{simple_multitask}
\end{figure*}

\begin{figure*}[h!]
\captionsetup[subfigure]{labelformat=empty}
\makebox[\textwidth][c]{
\subfloat[]{
  \includegraphics[width=0.4\linewidth]{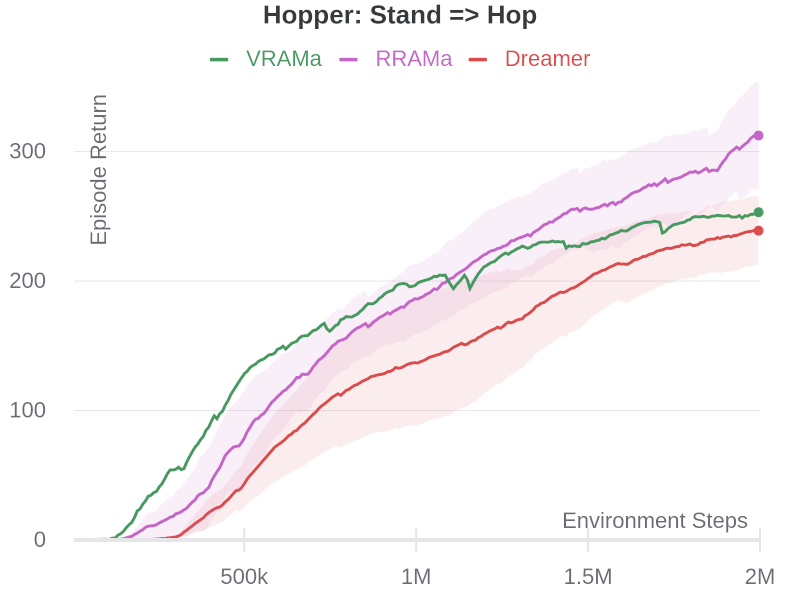}
}
\subfloat[]{
  \includegraphics[width=0.4\linewidth]{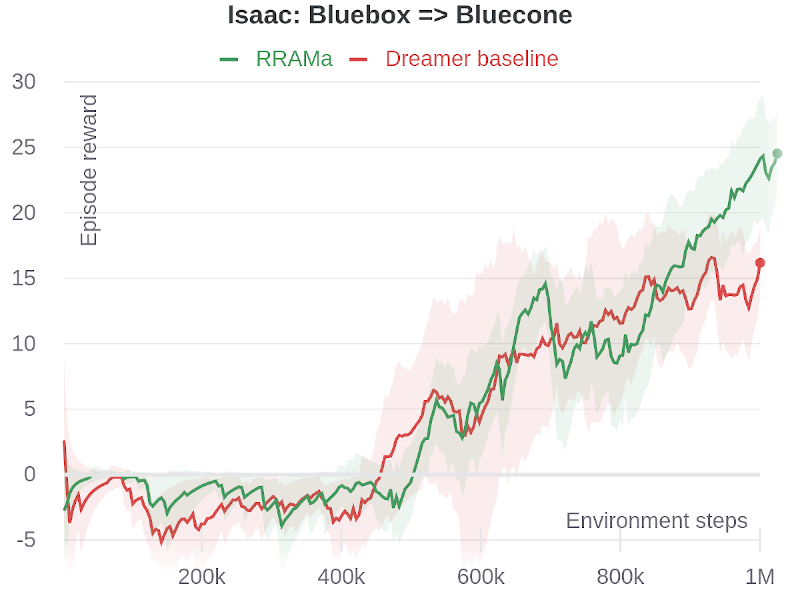}
}}
\vspace{-10mm}
\makebox[\textwidth][c]{
\subfloat[]{
  \includegraphics[width=0.4\linewidth]{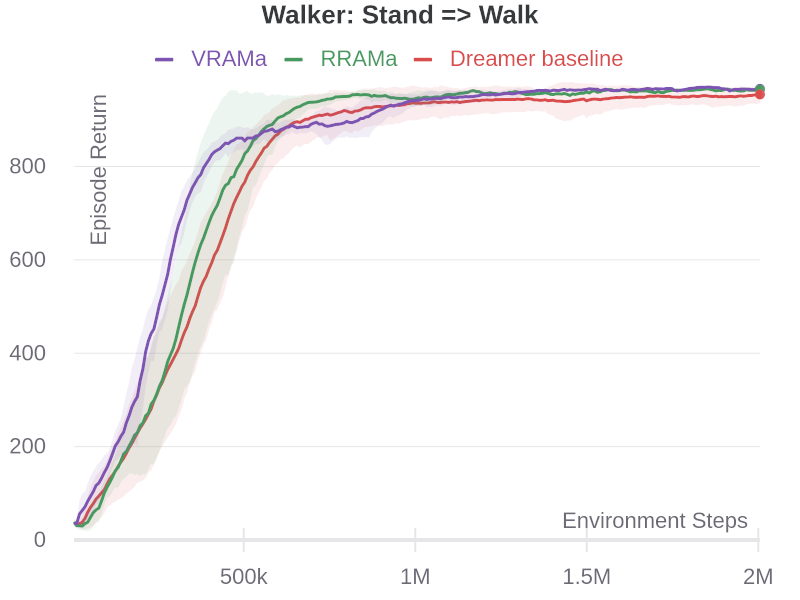}
}
\subfloat[]{
  \includegraphics[width=0.4\linewidth]{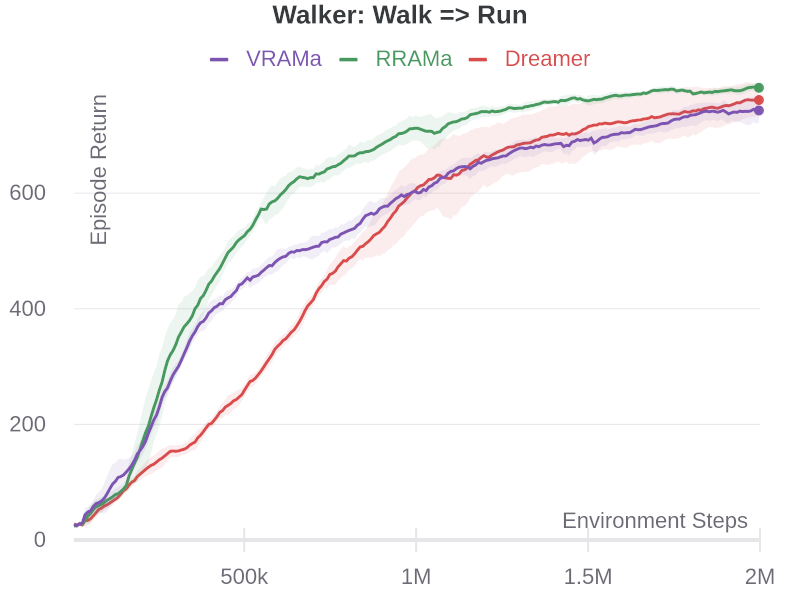}
}}

\vspace{10mm}
\caption{Performance plots for multitask experiment. We filled the $\mathcal{M}$ with one task experience: the Stand task for the Hopper and the Bluebox task for the Isaac. Next, we train the MBRL agents on different task having the prior experience in $\mathcal{M}$ }
\label{perf1}
\end{figure*}

We provide a detailed picture of how the model is performing on each reported environment. We conducted three lines of experiments. First, the experiments with the same task in the multitask buffer in DMC and Isaac environments are shown in Figure \ref{same_task_f}. Second, the results for the simple multitask case with Metaworld environment are shown in Figure  \ref{simple_multitask}. Finally, the full multitask experiments in DMC and Isaac environments are shown in Figure \ref{perf1}


\section{Ablation Study}
\label{ablation}

\begin{figure*}[h]
\captionsetup[subfigure]{labelformat=empty}
\makebox[\textwidth][c]{
\subfloat[]{
  \includegraphics[width=0.5\linewidth]{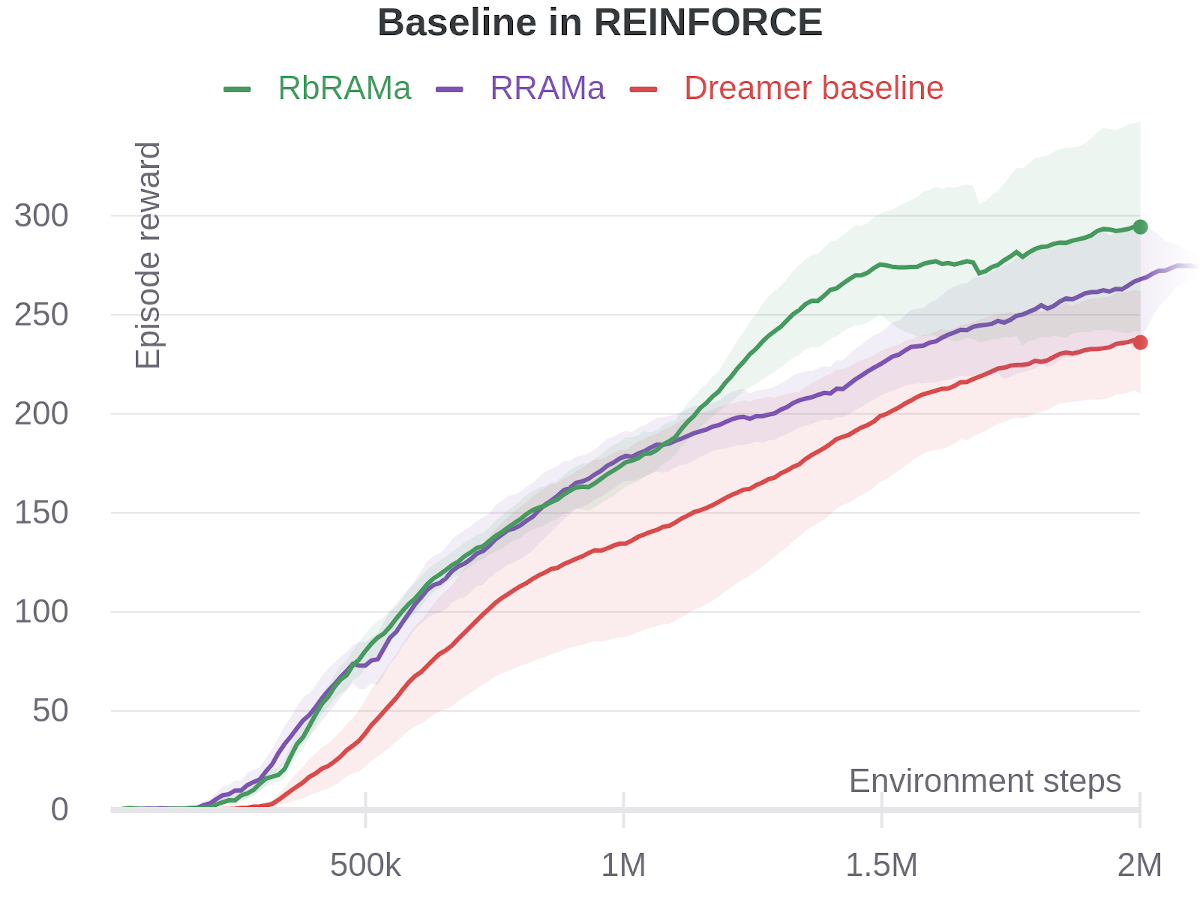}
}
\subfloat[]{
  \includegraphics[width=0.5\linewidth]{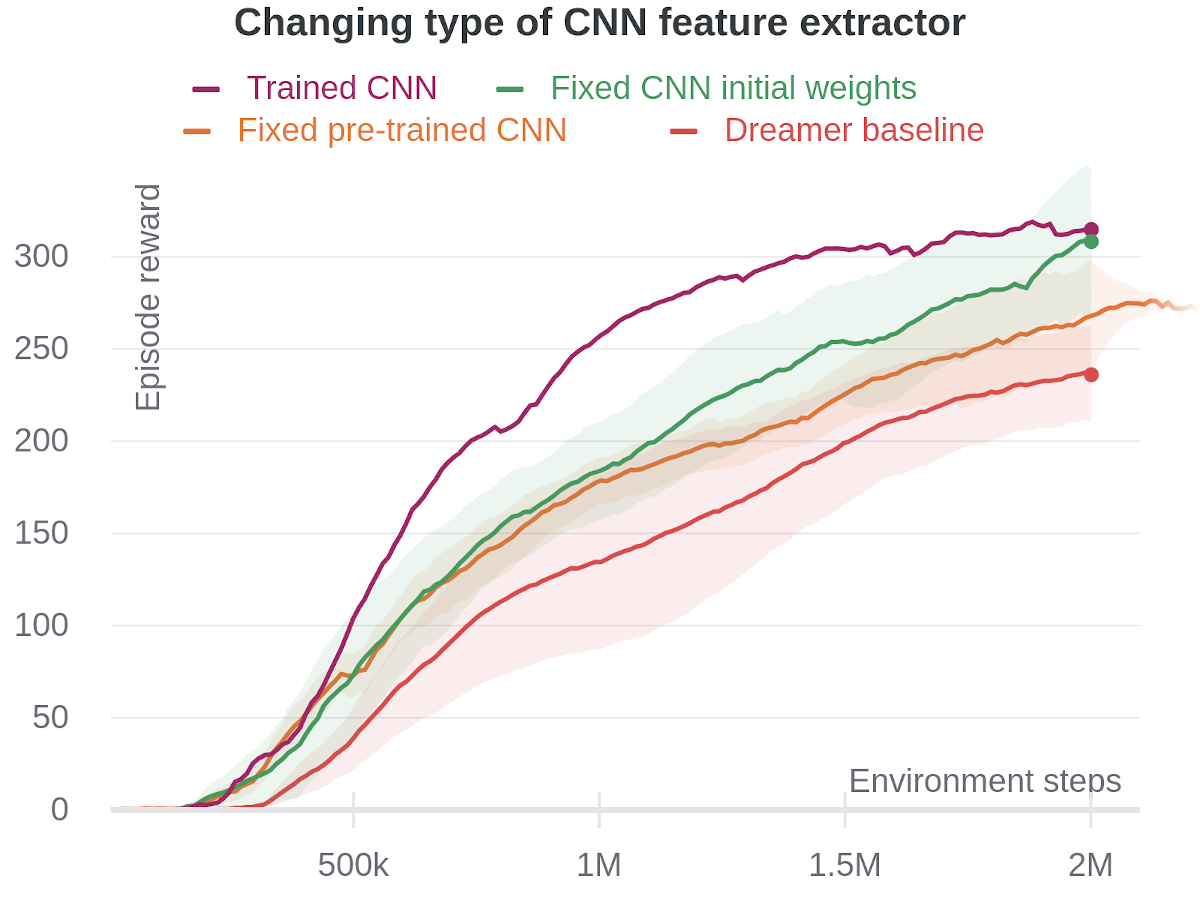}
}

}
\makebox[\textwidth][c]{
\subfloat[]{
  \includegraphics[width=0.5\linewidth]{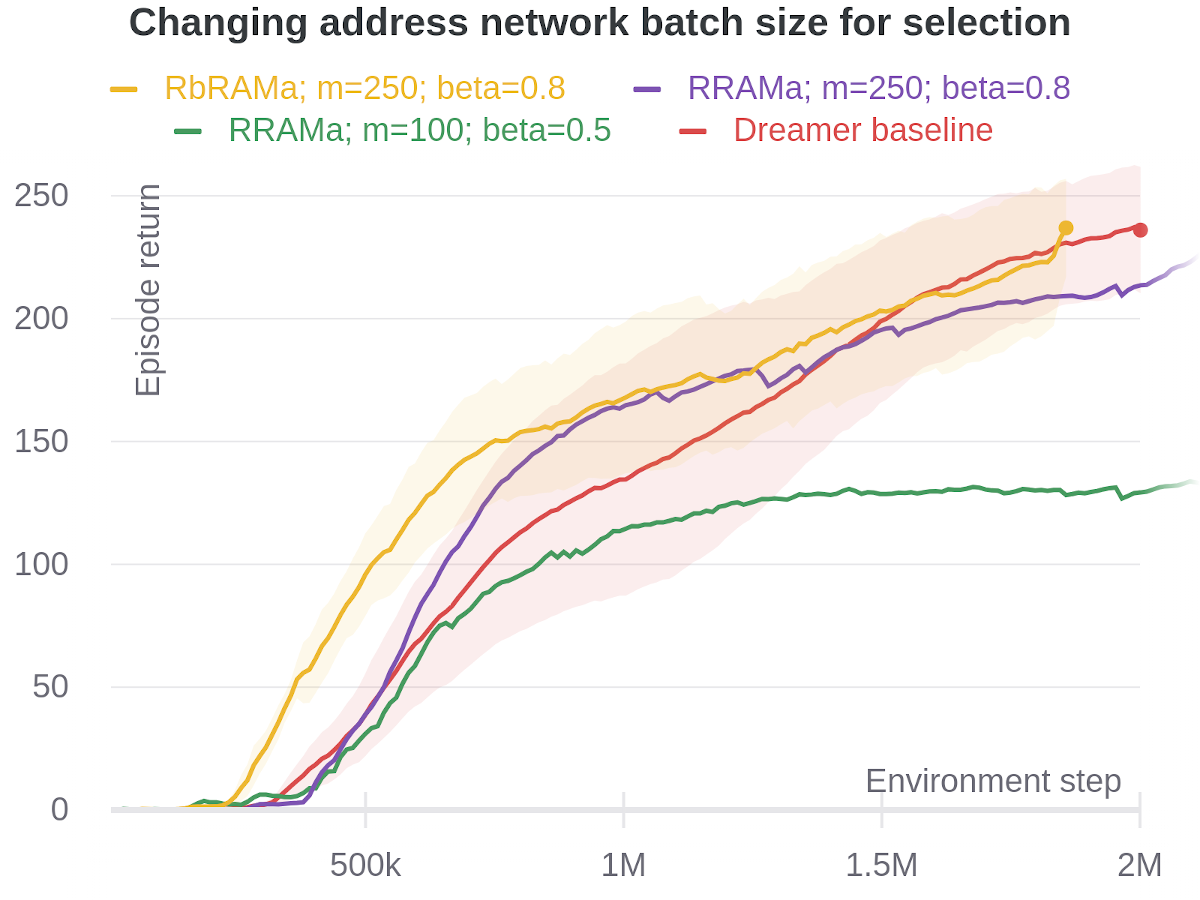}
}
}
\caption{Different hyperparameters for DeepMind Control domain}
\label{abl_dmc}
\end{figure*}

To choose better hyperparameters we ran short ablation study during which we optimized the choice of hyperparameters. We ran the experiments in the DMC and  we selected three primary directions for optimization: building better addressing objectives; optimizing representations for the addressing model; and increasing the diversity of the data inside batch for addressing model. In Figure \ref{abl_dmc} presented different runs for each hyperparameter. All hyperparameters were tested in Hopper Hop task with Hopper Stand experience in multitask storage.


\end{document}